\providecommand{\arxivversion}{1}
\documentclass[webpdf,contemporary,large,namedate]{oup-authoring-template}
\newcommand{\societylogo}{}

\ifnum\arxivversion=1
\makeatletter
\def\ps@opening{\let\@oddhead\@empty\let\@evenhead\@empty
  \def\@oddfoot{\hfil\sffamily\fontsize{8bp}{10bp}\selectfont\thepage\hfil}\let\@evenfoot\@oddfoot}
\let\ps@headings\ps@opening
\makeatother
\fi

\usepackage[T1]{fontenc}
\DeclareTextCommandDefault{\k}{\UseTextAccent{T1}\k}

\usepackage{mathtools}
\usepackage{bm}
\usetikzlibrary{decorations.pathreplacing, arrows.meta, calc, intersections}

\definecolor{metablue}{HTML}{800020}
\definecolor{metafg}{HTML}{1C2B33}
\definecolor{metabg}{HTML}{F1F4F7}
\definecolor{metagray}{RGB}{101,103,107}
\definecolor{theorembg}{HTML}{F2E6E4}
\definecolor{quotebg}{HTML}{D6E4F0}

\makeatletter
\renewcommand\paragraph{\@startsection{paragraph}{4}{\z@}%
  {1.5ex \@plus .5ex \@minus .2ex}{-1em}{\normalfont\normalsize\bfseries}}
\makeatother

\theoremstyle{thmstyleone}

\theoremstyle{thmstyletwo}

\theoremstyle{thmstylethree}
\newtheorem{definition}{Definition}

\newtheorem{hypothesis}{Hypothesis}

\newcommand{\calA}{\mathcal{A}}

\DeclareMathOperator*{\tr}{tr}
\DeclareMathOperator*{\softmax}{softmax}

\DeclareMathOperator*{\argmax}{arg\,max}
\DeclareMathOperator{\diag}{diag}

\DeclareMathOperator{\erk}{\rho}

\DeclareMathOperator{\relu}{relu}

\DeclareMathOperator{\integ}{\iota}
\DeclareMathOperator{\rk}{rk}
\newcommand{\bA}{\mathbf{A}}
\newcommand{\bB}{\mathbf{B}}
\newcommand{\bU}{\mathbf{U}}
\newcommand{\bD}{\mathbf{D}}
\newcommand{\bG}{\mathbf{G}}

\newcommand{\bI}{\mathbf{I}}
\newcommand{\bJ}{\mathbf{J}}
\newcommand{\bL}{\mathbf{L}}

\newcommand{\bP}{\mathbf{P}}

\newcommand{\bV}{\mathbf{V}}
\newcommand{\bzero}{\boldsymbol{0}}

\newcommand{\bSigma}{\boldsymbol{\Sigma}}
\newcommand{\bW}{\mathbf{W}}
\newcommand{\bR}{\mathbf{R}}
\newcommand{\ba}{\mathbf{a}}

\newcommand{\bb}{\mathbf{b}}

\newcommand{\bg}{\mathbf{g}}
\newcommand{\bh}{\mathbf{h}}

\newcommand{\bs}{\mathbf{s}}
\newcommand{\bS}{\mathbf{S}}
\newcommand{\bu}{\mathbf{u}}
\newcommand{\bk}{\mathbf{k}}

\newcommand{\bv}{\mathbf{v}}
\newcommand{\bw}{\mathbf{w}}
\newcommand{\bx}{\mathbf{x}}
\newcommand{\bX}{\mathbf{X}}
\newcommand{\by}{\mathbf{y}}
\newcommand{\bz}{\mathbf{z}}

\newcommand{\cO}{\mathcal{O}}

\newcommand{\cV}{\mathcal{V}}
\newcommand{\cW}{\mathcal{W}}

\newcommand{\pending}[1]{\textcolor{red}{[#1]}}

\colorlet{teallt}{teal!30}      %
\colorlet{teallt2}{teal!15}     %
\colorlet{tealmid}{teal!60}     %
\colorlet{ink}{metafg}          %
\colorlet{crimson}{metablue}    %

\providecommand{\tr}{\operatorname{tr}}
\providecommand{\bB}{\mathbf{B}}
\providecommand{\bzero}{\mathbf{0}}

\newcommand{\panel}[2]{%
  \begin{tikzpicture}
    \node[anchor=south west, inner sep=0] (img) {#2};
    \node[anchor=north west, font=\bfseries, inner sep=2pt]
         at (img.north west) {#1};
  \end{tikzpicture}%
}

\newcommand{\secref}[1]{\S\ref{#1}}

\begin{document}

\journaltitle{Neuroscience of Consciousness}
\DOI{DOI added during production}
\copyrightyear{YEAR}
\pubyear{YEAR}
\vol{XX}
\issue{x}
\access{Published: Date added during production}
\appnotes{Research Article}

\firstpage{1}

\title[Differential functionalism]{Differential Functionalism: Jacobians Shape\\ the Structure of Phenomenal Experience}

\ifnum\arxivversion=1
\author[$\ast$]{David Balduzzi}
\corresp[$\ast$]{\href{mailto:dbalduzzi@gmail.com}{dbalduzzi@gmail.com}}
\else
\author[1,$\ast$]{David Balduzzi}
\address[1]{\orgname{\pending{affiliation}}}
\corresp[$\ast$]{Corresponding author. \href{mailto:dbalduzzi@gmail.com}{dbalduzzi@gmail.com}}
\fi

\received{Date}{0}{Year}
\revised{Date}{0}{Year}
\accepted{Date}{0}{Year}

\abstract{This paper investigates the hypothesis that the first-order structure of physical interactions (Jacobians) shapes the structure of phenomenal experience.
  Two measures of Jacobian structure are introduced: transparency, the effective rank, which counts the independent differences that make a difference; and integrity, the normalised algebraic connectivity, which measures how hard a system is to split. Applied to a series of worked examples, the framework yields correlates of:
  (1) experiences that overlap without systems merging;
  (2) the boundaries of a system, which are selected by focus and limited by noise;
  (3) temporal thickness, the persistence of the past into the present, lengthened by a spread of timescales;
  (4) spatial structure under convolution, where unity at scale needs long-range connections;
  (5) the distinction between what is vivid and what is obscure, for errors and attention heads;
  (6) the incoherence of a randomly initialised network, as a structural model of James's ``blooming, buzzing confusion'';
  (7) distinct versus confused ideas;
  (8) learning as accumulated experience; and, conjecturally,
  (9) the contrast between memory as duration in brains and as storage in computer hardware.
  The measures can be computed at scale in artificial networks and estimated from perturbational data in brains.}

\keywords{phenomenal structure; Jacobians; effective rank; algebraic connectivity; integrated information; neural networks}

\maketitle

\section{Introduction}
\label{sec:intro}

\noindent
Consciousness science has well-developed tools for asking \emph{whether} a system is conscious, from contrastive studies of neural correlates to perturbational indices that separate waking from sleep and anaesthesia \citep{massimini2005,casali2013,alkire2008}. It has fewer tools for asking how experience is \emph{structured}: how it is bounded, how far it extends in time, which of its parts are vivid and which obscure. Recent work proposes to characterise experiences by the relations among them, and to look for matching relations in physical systems \citep{tsuchiya2022,kleiner2024,oizumi2026}. Connecting physical structure to experience requires measures that apply to any system with a differentiable model, that can be computed at scale, and that can be estimated from data. Integrated information theory (IIT) offers a well-developed candidate \citep{balduzzi2008,oizumi2014,albantakis2023}, but its exact computation searches over all states and partitions of a system which confines it to toy examples.

This paper proposes an alternative approach, \emph{differential functionalism}, based on Jacobians, matrices of first-order sensitivities of a system's outputs to its inputs and parameters. Estimates of Jacobians can be constructed by techniques from perturbational neuroscience, stimulating one site and recording the responses at others \citep{massimini2005,casali2013}, or effective-connectivity models \citep{friston2003}. Neural nets are trained using Jacobians computed by backprop, which matters as speculation grows about whether they are conscious \citep{aru2023,butlin2023,chalmers2023}.

Jacobians are useful windows into differentiable systems. They compose by the chain rule, so the structure can be traced from one part to the next. They are evaluated at the system's actual state rather than over all the states it might have been in, making them cheaper than information-theoretic approaches based on counterfactuals. 

\subsection{The hypothesis}
\label{ssec:hypothesis}

Ernst Mach proposed a principle of ``complete parallelism'' between the psychical and the physical: a sensation cannot vary unless its immediate physical condition varies, and vice versa \citep{mach1914}. Parallelism is a claim about covariation, which Jacobians describe to first order. Hypothesis~\ref{h:first} is a principle of first-order parallelism: it ties the structure of experience to the Jacobians of the interactions that compose it.

\renewcommand{\thehypothesis}{G}
\begin{hypothesis}
  \label{h:first}
  The first-order structure of physical interactions, Jacobians, shapes the structure of phenomenal experience. Experience registers differences to the extent that they make a difference (\textbf{difference}), and extends to the extent that interactions connect (\textbf{connection}).
\end{hypothesis}

\noindent
\paragraph{Difference.}
Experience is transparent: we see \emph{through} air, as well as our eyes and neurons, act through limbs, and feel through tools. Transparency comes in degrees. The chain rule traces a perturbation through nested transformations and says how far it propagates before it is attenuated, blocked or absorbed. A difference that fails to propagate makes no difference, and by Hypothesis~\ref{h:first} does not contribute to experience. \emph{Transparency}, the effective rank of a Jacobian, counts the effective number of independent directions in which perturbations make a difference, \secref{ssec:analysis}.

\paragraph{Connection.}
Experience extends over time and across space. Some kind of glue binds the experience of interactions together. The glue can only be the structure of the interactions themselves. \emph{Integrity}, the normalised algebraic connectivity of the Jacobian's graph, measures how hard a system is to split: it is low when a weak seam nearly divides it and zero when it decomposes into noninteracting subsystems, \secref{ssec:analysis}. Together with transparency it can be used to determine boundaries of systems, \secref{ssec:boundaries}, and how far a system's past affects its present, \secref{ssec:time}.

\paragraph{Failure modes.}
A more general form of the hypothesis is that \emph{the structure of experience is shaped by the structure of physical interactions}, which is hard to test because it does not say which structure. This paper commits to first order. The difference clause fails if experience changes under a perturbation that the Jacobian maps to zero. The connection clause fails if a unified experience spans parts that no first-order interaction connects: for instance, if unified report across the hemispheres survived interruption of every interhemispheric pathway. The measures would fail if, for instance, experience across waking, dreaming sleep and anaesthesia does not rise and fall with integral rank (\secref{ssec:analysis}). Because the relevant Jacobians are closely related to estimates resulting from intracranial stimulation and TMS-EEG, these conditions can be checked, \secref{app:tests}.

\subsection{Overview}
\label{ssec:overview}

\S\ref{sec:gradland} introduces transparency and integrity and a provisional rule for selecting a system's boundaries based on them: a unit is a local maximum of integral rank. \S\ref{sec:life} presents worked examples in artificial networks, and draws out what each may imply for brains.

\paragraph{Contributions.}
\begin{enumerate}
  \item \emph{Two measures of Jacobian structure}, transparency and integrity, computable from singular values without searching over partitions or states. In brains they can be estimated from the response matrices that perturbational studies record, \secref{app:tests}.
  \item \emph{Boundaries.} Shared inputs make experiences overlap without making systems merge, \secref{ssec:world_in_me}; focusing selects where a perceiving system's boundary falls and noise limits how far it extends, \secref{ssec:boundaries}.
  \item \emph{Duration: temporal structure.} Mixing across time over a single timescale selects an instantaneous present; a spread of timescales, as found along the cortical hierarchy \citep{murray2014,hasson2008}, selects an extended one, \secref{ssec:time}.
  \item \emph{Spatial structure.} For convolutions, transparency captures how evenly the kernel passes spatial frequencies, and integrity how well it mixes the slowest spatial mode. The analysis also suggests local wiring is not sufficient for high integrity in large systems; long-range connections may be needed, \secref{ssec:space}.
  \item \emph{Structure within experience.} The relationship between the input and facets of the Jacobian can be computed and interpreted in relu nets, \secref{ssec:facets}. Jacobians can distinguish the vivid from the obscure, in errors and attention, \secref{ssec:loss}--\ref{ssec:attention}; give insight into when a system lacks the invariants needed to parse its inputs, \secref{ssec:init}; and separate distinct from confused interactions, \secref{ssec:distinct}.
  \item \emph{Learning from experience.} Weight updates have Jacobians of their own, so learning adds second-order structure. A net's knowledge is its accumulated error gradients, i.e. facets of its experience, \secref{ssec:learn}.
  \item \emph{Substrate dependence.} Because the hypothesis concerns physical interactions, implementation details are consequential: stored memory and tiled computation fragment interactions, \secref{ssec:hardware}.
\end{enumerate}

\subsection{Related work: why gradients?}
\label{ssec:gradients}

The approach is related to IIT \citep{balduzzi2008,oizumi2014,albantakis2023}, and shares its starting point that a system which decomposes into independent pieces is not one system. Transparency and integrity play roles analogous to effective and integrated information in IIT. The main difference is using gradients rather than information theory. Integrated information is defined over all possible states and partitions of a system, taking $\cO(e^n)$ time for $n$ neurons. Given an $n\times d$ Jacobian, the measures below take $\cO(nd\cdot\min(n,d))$ time using a singular value decomposition (SVD). Integrated information has been computed exactly only for systems of a dozen or so neurons \citep{mayner2018}.

Differential functionalism and IIT disagree, for example on whether experiences can overlap, \secref{ssec:world_in_me}, and on feedforward networks, which can have nonzero integral rank but always have zero integrated information \citep{oizumi2014}; by the unfolding argument \citep{doerig2019} a recurrent network is feedforward over time, \secref{ssec:time}. Differential functionalism applies to systems and setups out of IIT's reach, including trained convolutional nets, attention heads and networks at initialisation, \secref{ssec:space}--\ref{ssec:init}, and to learning, \secref{ssec:learn}.

Gradients and information are loosely related. An AND-gate that outputs $1$ rules out more alternatives than one that outputs $0$; relatedly, the path to $1$ is more sensitive to perturbations of its inputs. Sensitivity to perturbation also underlies the perturbational complexity index \citep{casali2013}.

Differential functionalism makes contact with several theories of consciousness, for reviews see \citet{seth2022a,butlin2023,kuhn2024,mudrik2025}. Global workspace theory \citep{baars1988} holds that states are conscious when they enter a workspace that broadcasts across specialised modules; a workspace, if the term picks out a unit, should be a local maximum of integral rank, which can be checked in language models where Jacobians locate one \citep{gurnee2026}, \secref{app:tests}. Recurrent processing theory \citep{lamme2006} ties consciousness to recurrence; here duration needs mixing across time, \secref{ssec:time}. Learning adds second-order structure, a point of contact with higher-order theories \citep{rosenthal2006,lau2008,cleeremans2011}, \secref{ssec:learn}.

\section{Methods}
\label{sec:gradland}

This section introduces notation and provides tools for analysing Jacobians in idealised models: compositions of matmuls and other (mostly) differentiable operations, whose Jacobians are known exactly. \secref{ssec:hardware} considers some aspects of hardware implementation.

Loss functions and their gradients are not privileged below, although they are important examples, \S\ref{ssec:loss}, and play a larger role in learning, \S\ref{ssec:learn}.

\subsection{Notation and basics}
\label{ssec:ops}

There is a directed acyclic graph (DAG) $G$. Nodes of $G$ are marked by spacetime coordinates $(j, t)$. Spatial coordinates $j$ refer to neurons. Temporal coordinates $t$ index activity at different times. Each neuron has a parameter vector $\bw_{(j,t)}$, which can change with time but won't until we consider learning in \S\ref{ssec:learn}, and an activation function $f_j:{\mathbb R}\rightarrow {\mathbb R}$. The output at $(j, t)$ is $f_j\left(\bw_{(j,t)}^\intercal \cdot \bx_{\bullet\rightarrow(j,t)}\right)$ where $\bx_{\bullet\rightarrow(j,t)}$ collects neuron $j$'s inputs at time $t$ into a vector.

Given a graph $G$, define a \textbf{system} as a subgraph $S$, with some nodes of $S$ labelled as inputs and outputs. The inputs and outputs of $S$ are nonempty and disjoint. Examples of outputs are: the loss function; the outputs of a particular layer; or all outputs of all layers except the first (assuming it is the input). The parameters of input nodes are ignored --- as if they do not exist. Let $\cW$ denote all parameters in a system.

The specification of inputs and outputs matters. A pathological example is to put outputs \emph{upstream} of inputs. The Jacobian is the zero matrix; the system is inert and the measures in \S\ref{ssec:analysis} return zero.

Three main neural architectures are important:
\begin{enumerate}
  \item \emph{Multi-layer perceptron (MLP).}
  There are $L+1$ layers of $\{d_l\}_{l=0}^L$ neurons. There is a connection from every neuron in layer $l$ to every neuron in layer $l+1$. The spatial coordinate now has two dimensions: the layer and position in the layer. Assuming $T$ time points, the graph $G(\text{MLP})$ is $T$ disconnected subgraphs. Each subgraph is a copy of the diagram of the MLP. Subgraphs are disconnected because there is no interaction across time (again, there is no learning for now).

  Suppose all neurons in an MLP have activation function $f:{\mathbb R}\rightarrow {\mathbb R}$. Standard notation for the computation performed by each layer is then $\bz^{(l)} = f(\ba^{(l)})$ where $\ba^{(l)} = \bW_l \bz^{(l-1)}$.
  \item \emph{Recurrent neural network (RNN).}
  An RNN has feedforward and recurrent layers. The recurrent updates look like
  \begin{equation}
    \label{eq:rnn}
    \bh_{t+1} = f_\bw(\bh_t, \bx_{t+1})
  \end{equation}
  for some function $f$.
  \item \emph{Transformer.}
  See standard references for description of transformers \citep{vaswani2017,chowdhery2022}. This paper only considers a simplified model of attention, see \S\ref{ssec:attention}. 
\end{enumerate}

\paragraph{Modelling choices.}
Applying the measures requires a model, which involves various choices.

The first is the \emph{grain}: which variables count as nodes, and at what spatial and temporal resolution. The grain is chosen to suit the phenomena of interest. A model of subatomic particles is unhelpful for understanding human behaviour; a model of neurons, or of the elements of an artificial network, is appropriate. Throughout, the grain is the neuron of an artificial network, the pixel of a grid, \secref{ssec:space}, or the step of a recurrent update, \secref{ssec:time}.

The second is which variables are parameters and state. Physically there is only state since cells --- and chips --- change continually. A variable acts as a parameter over some time frame to the extent that it is stable across it. Synaptic weights are parameters over milliseconds and state over a lifetime. In neural nets weights are fixed during the forward pass, and are modified during learning, \secref{ssec:learn}. Values held in computer memory are discussed in \secref{ssec:hardware}.

The third is, given a model, which nodes form the system: its outputs, parameters, and inputs. These are not free: Definition~\ref{def:boundary} selects them by integral rank, though a modeller can fix them by hand, for instance to compare systems over a common window.

\subsection{Quantifying Jacobian structure}
\label{ssec:analysis}

There are two concepts to make precise to operationalise Hypothesis~\ref{h:first}. \emph{Transparency} quantifies the difference clause: how many independent differences make a difference, and how evenly. It vanishes exactly when nothing makes a difference. \emph{Integrity} quantifies the connection clause: how strongly interactions bind a system. It vanishes exactly when they fail to connect it.

\paragraph{Input and parameter Jacobians.}
Given a system of $d$ inputs $\bx$, neurons with $p$ parameters $\cW$, and $n$ outputs $\by$ consider the $n\times d$ \emph{input Jacobian} and the $n\times p$ \emph{parameter Jacobian}
\begin{equation*}
  \bJ_{\by\leftarrow \bx} = \nabla_{\bx} \by(\bx, \cW)
  \quad\text{and}\quad
  \bJ_{\by\leftarrow \cW} = \nabla_{\cW} \by(\bx, \cW)
\end{equation*}
The two matrices could be combined; it is convenient to keep them separate since they play different roles. $\bJ_{\by\leftarrow \bx}$ is the proposed correlate of perceiving and $\bJ_{\by\leftarrow \cW}$ of choosing. The input Jacobian shapes how the system views the world: which aspects it is sensitive to, or not. The parameter Jacobian captures affordances of a sort \citep{gibson1979}: the sensitivity of the levers the system pulls to modify its behaviour. In neuroscience terms, the input Jacobian is a linear response matrix: what perturbation experiments measure and effective-connectivity models estimate, \secref{app:tests}.

Jacobians have a rich structure. The sets of inputs and outputs can be flexibly chosen. They could be activity across multiple layers, or neuronal activity at different time points in a sequence, or both. The parameter Jacobian inherits structure from the architecture itself: interactions between layers, blocks and subsystems can be extracted as needed.

\begin{figure}[ht]
  \centering
  \resizebox{0.95\columnwidth}{!}{\newcommand{\zone}[4]{%
  \filldraw[fill=teal!7, draw=teal!42, line width=0.5pt, rounded corners=2pt]
    (#1,#3) rectangle (#2,#4);
}

\definecolor{tray1}{HTML}{0B4F4A}
\definecolor{tray2}{HTML}{2D8A80}
\definecolor{tray3}{HTML}{74B8AF}
\definecolor{tray4}{HTML}{C2E2DC}

\newcommand{\raypanel}[6]{%
  \begin{scope}[xshift=#1 cm, yshift=#2 cm]
    \draw[tray1] (0,1.1)  -- (1.0,1.1)  to[out=0,in=180] (1.8,#3) -- (2.8,#3);
    \draw[tray2] (0,0.5)  -- (1.0,0.5)  to[out=0,in=180] (1.8,#4) -- (2.8,#4);
    \draw[tray3] (0,-0.1) -- (1.0,-0.1) to[out=0,in=180] (1.8,#5) -- (2.8,#5);
    \draw[tray4] (0,-0.7) -- (1.0,-0.7) to[out=0,in=180] (1.8,#6) -- (2.8,#6);
  \end{scope}
}

\begin{tikzpicture}[line width=0.85pt]

\node[font=\normalsize\bfseries, anchor=south] at (1.4,1.55) {Low $\erk(\bm{A})$};
\zone{1.0}{1.8}{-0.95}{1.35}
\raypanel{0}{0}{0.30}{0.22}{0.14}{0.06}

\node[font=\normalsize\bfseries, anchor=south] at (5.4,1.55) {High $\erk(\bm{A})$};
\begin{scope}[xshift=4.0cm]
  \zone{1.0}{1.8}{-0.95}{1.35}
\end{scope}
\raypanel{4.0}{0}{0.5}{1.0}{0.1}{-0.7}

\begin{scope}[yshift=-3.3cm]

  \node[font=\normalsize\bfseries, anchor=south] at (1.4,1.55) {$\integ(\bm{A})=0$};
  \begin{scope}
    \zone{1.0}{1.8}{-0.95}{1.35}
    \draw[gray!55, densely dashed, line width=0.5pt] (1.0,0.2) -- (1.8,0.2);
  \end{scope}
  \raypanel{0}{0}{0.6}{0.9}{-0.5}{-0.2}

  \node[font=\normalsize\bfseries, anchor=south] at (5.4,1.55) {High $\integ(\bm{A})$};
  \begin{scope}[xshift=4.0cm]
    \zone{1.0}{1.8}{-0.95}{1.35}
  \end{scope}
  \raypanel{4.0}{0}{-0.65}{-0.1}{0.5}{1.05}

\end{scope}

\end{tikzpicture}}
  \caption{Matrices stylistically depicted as lenses. \textbf{Top:} transparency. Low $\erk(\bm{A})$: rays converge to nearly one output, a single dominant impression; high $\erk(\bm{A})$: many distinct impressions pass through. \textbf{Bottom:} integrity. $\integ(\bm{A})=0$: some rays do not interact; high $\integ(\bm{A})$: rays are tangled.}
  \label{fig:transparency_cohesion}
  \figalttext[Matrices drawn as lenses, illustrating transparency and integrity.]{Four lens diagrams. Top row: rays passing through a low-transparency matrix converge onto nearly one output, while through a high-transparency matrix many distinct rays pass. Bottom row: with zero integrity some rays pass through without interacting; with high integrity the rays are tangled together.}%
\end{figure}

\paragraph{Transparency.}
Think of Jacobians as akin to lenses, Fig.~\ref{fig:transparency_cohesion}. How much can you see through them? The singular value decomposition (SVD) factors a matrix into \emph{(1)} rotation to latent coordinates; \emph{(2)} coordinatewise rescaling; and \emph{(3)} rotation to output coordinates, see \S\ref{app:svd}. The singular values determine whether a matrix transmits many independent directions or few. If all singular values are zero then the matrix is completely opaque: nothing passes through it. If one singular value is large and the rest are close to zero, then one direction is transmitted strongly and everything else (in the latent coordinates) is attenuated or zeroed out.

\begin{definition}
  \label{def:erk}
  Let $\bA_{n\times d}$ be a matrix with rank $r$, and nonzero singular values $\sigma_1,\ldots, \sigma_r$.
  The \textbf{transparency} or \textbf{effective rank} of $\bA$ is
  \begin{equation*}
    \erk(\bA) := \begin{cases}
      \frac{\left(\sum_{i=1}^{r}\sigma_i^2\right)^2}{\sum_{i=1}^r \sigma_i^4} & \text{when } \rk(\bA) > 0 \\
      0 & \text{else}
   \end{cases}
  \end{equation*}
\end{definition}
\noindent
Effective rank is known as the \emph{participation ratio} in condensed matter physics \citep{bell1970}. It is called the \emph{dimension of a representation} in the neuroscience literature \citep{gao2015,litwin-kumar2017}, see \S\ref{app:erk}, where it is computed on covariance matrices of neural activity. Here, in contrast, the object is the Jacobian, which captures state-dependent transmission of differences.  It satisfies:
\begin{itemize}
  \item \emph{Scale invariance:}
  $\erk(\alpha\cdot \bA) = \erk(\bA)$ for $\alpha>0$.
  \item \emph{Permutation invariance:}
  $\erk(\bP_n\bA\bP_d)=\erk(\bA)$ for permutations $\bP_n$ and $\bP_d$.
  \item \emph{Bounds:}
  $0\leq \erk(\bA) \leq \rk(\bA)$, with
  $\erk(\bA)=\rk(\bA)$ iff all nonzero singular values are equal.
  \item \emph{Zero condition:}
  $\erk(\bA) = 0$ iff $\bA=0$.
  \item \emph{Maximal condition:} Effective rank is maximised when all singular values are nonzero and equal, \textbf{when the number of independent directions transmitted is maximised and they are transmitted equally strongly}, see Fig.~\ref{fig:erk_kc}.
\end{itemize}

\noindent
If $\erk(\bA)=0$ then $\bA$ is completely opaque, as a view, or inert, as a bundle of affordances. This happens, for example, when a system does not fit in a light cone: if the input is light emitted by the sun a minute ago and the output is light hitting the earth now, then effective rank is zero, since light does not travel fast enough for perturbations to those inputs to affect these outputs. Such a system is inert.

\paragraph{Integrity.}
How does a Jacobian hang together? Is it one or many, knit or split? Do signals pass through in parallel, or do they interact?

An $n\times d$ matrix $\bA$ acts via $n$ dot products:
\begin{equation*}
  \mathbf{A}\bx = \left(\begin{array}{c}
    \mathbf{a}_1^\intercal \bx\\
    \vdots\\
    \mathbf{a}_n^\intercal \bx\\
  \end{array}\right)
  \quad\text{where $\mathbf{a}^\intercal_i$ are the rows of $\mathbf{A}$}.
\end{equation*}
The $n$ dot-products are \emph{not} $n$ independent computations in general, because they share input $\bx$. If $\mathbf{A}$ is block-diagonal
\begin{equation*}
  \mathbf{A}\bx =
  \left(\begin{array}{cc}
    \mathbf{A}_{11} & \mathbf{0} \\
    \mathbf{0}  & \mathbf{A}_{22}
  \end{array}\right)
  \left(\begin{array}{c}
    \bx_1 \\
    \bx_2
  \end{array}\right)
  = \left(\begin{array}{c}
    \mathbf{A}_{11}\bx_1\\
    \mathbf{A}_{22}\bx_2
  \end{array}\right)
\end{equation*}
then computations in the blocks are independent. If $\bA$ is diagonal, channels do not mix with one another, although each nonzero diagonal entry transmits a difference. The decomposition into blocks is basis-dependent. That is as it should be: Jacobians have a \emph{physically meaningful} basis specified by their input and output nodes.

Exact block structure is the extreme case. More often a system has weak seams between parts. Construct the weighted, undirected bipartite graph $G(\bA)$ with $d$ input nodes and $n$ output nodes, and an edge of weight $|A_{ij}|$ between input $j$ and output $i$. Remove identically zero rows and columns to obtain $\bA_*$ and its \emph{active graph} $G_*(\bA)=G(\bA_*)$, with $n_*$ outputs and $d_*$ inputs. For $\bA\ne0$, rescale $|\bA_*|$ by its positive row and column sums,
\begin{equation*}
  \calA:=\bD_r^{-1/2}\cdot|\bA_*|\cdot\bD_c^{-1/2},
\end{equation*}
where $\bD_r=\diag(|\bA_*|\cdot\mathbf{1})$ and $\bD_c=\diag(|\bA_*|^\intercal\cdot\mathbf{1})$. For a nonzero matrix the largest singular value of $\calA$ is $1$.

\begin{definition}
  \label{def:integrity}
  For $\bA\ne0$, the \textbf{integrity} of $\bA$ is $\integ(\bA):=1-\sigma_2(\calA)$, where $\sigma_2$ is the second-largest singular value when $\min(n_*,d_*)>1$, and $\sigma_2(\calA):=0$ when $\min(n_*,d_*)=1$. Set $\integ(\bA):=0$ for $\bA=0$. Inert nodes mean identically zero rows or columns of the Jacobian, and are removed before normalisation.
\end{definition}
\noindent
Integrity is the normalised algebraic connectivity of the active graph, capped at $1$ \citep{fiedler1973,chung1997}, which matters only for a $1\times1$ Jacobian, \secref{app:integrity}. Removing inert nodes means measuring the active circuit. It satisfies:
\begin{itemize}
  \item \emph{Scale invariance:} $\integ(\alpha\cdot\bA)=\integ(\bA)$ for $\alpha>0$.
  \item \emph{Permutation invariance:} $\integ(\bP_n\bA\bP_d)=\integ(\bA)$ for permutations $\bP_n$ and $\bP_d$.
  \item \emph{Bounds:} $0\le\integ(\bA)\le1$.
  \item \emph{Zero condition:} $\integ(\bA)=0$ iff $\bA=0$ or $G_*(\bA)$ is disconnected, i.e.\ the nonempty active matrix can be permuted into more than one nonzero diagonal block.
  \item \emph{Maximal condition:} $\integ(\bA)=1$ iff $|\bA|$ has rank one, $|A_{ij}|=u_iv_j$ for some vectors $\bu_n$ and $\bv_d$.
  \item \emph{Weakest seam:} by Cheeger's inequality, $\integ/2\le h\le\sqrt{2\integ}$, where $h$ is the minimum crossing weight divided by the smaller side's total degree \citep{chung1997}.
\end{itemize}

\noindent
For a nonzero Jacobian, integrity is zero iff its active input and output nodes split into disconnected groups. If what an animal sees never interacts with what it hears, integrity is zero and the animal has a kind of blind spot: it cannot make decisions based on combinations of sights and sounds. If sight and sound interact only through a thin channel, integrity is low.

Absolute values $|A_{ij}|$ are used so that interactions cannot cancel: an inhibitory influence connects as much as an excitatory one. Computing it takes one singular value of $\calA$.

\paragraph{Integral rank.}
Combine transparency and integrity to quantify both how much a system transmits and how tightly it hangs together:
\begin{definition}
  \label{def:integral}
  The \textbf{integral rank} of $\bA$ is $\integ_{\erk}(\bA) := \erk(\bA)\cdot\integ(\bA)$.
\end{definition}
\noindent
Integral rank is scale and permutation invariant, satisfies $0\le\integ_{\erk}(\bA)\le\erk(\bA)$, and vanishes iff $\bA$ is zero or its active graph is disconnected. If $\integ_{\erk}(\bA)=0$ then either nothing makes a difference, or the system decomposes into independent subsystems.

Transparency and integrity pull in different directions. The easiest way to maximise transparency is a nonzero multiple of the identity, which has zero integrity in dimensions greater than one. Dynamical isometry, where all singular values of the Jacobian are close to 1, is useful when initialising neural nets and yields near maximal transparency \citep{saxe2014,pennington2017,xiao2018,tarnowski2019}. The easiest way to maximise integrity is a matrix with all entries equal, which has transparency one. Some matrices score high on both. An $n\times n$ Hadamard matrix attains the maximum possible integral rank of $n$ (in dimensions where one exists). Hadamard matrices are orthogonal up to scale, with all entries in $\pm1$. \textbf{Integral rank is high for matrices that mix well: that pass many independent differences, and cannot be split.}

\begin{figure}[ht]
  \centering
  \begin{minipage}[c]{0.9\columnwidth}
    \centering
    \resizebox{\linewidth}{!}{\tikzset{ecell/.style={minimum size=1.1cm, inner sep=0pt, draw=teal!60, line width=0.42pt}}
\pgfmathsetmacro{\B}{1.1}\pgfmathsetmacro{\BH}{3*\B}
\pgfmathsetmacro{\bcl}{3*\B+0.62}\pgfmathsetmacro{\bw}{0.40}\pgfmathsetmacro{\bs}{0.62}
\begin{tikzpicture}
\begin{scope}[xshift=0.0cm]
  \node[font=\normalsize\bfseries, anchor=south] at ({(\bcl+2*\bs+\bw)/2},{3*\B+0.42}) {Low $\erk(\bm{A}) = 1.2$};
  \node[ecell, fill=teal!52] at ({0.5*\B},{2.5*\B}) {};
  \node[ecell, fill=teal!38] at ({1.5*\B},{2.5*\B}) {};
  \node[ecell, fill=teal!60] at ({2.5*\B},{2.5*\B}) {};
  \node[ecell, fill=teal!60] at ({0.5*\B},{1.5*\B}) {};
  \node[ecell, fill=teal!15] at ({1.5*\B},{1.5*\B}) {};
  \node[ecell, fill=teal!30] at ({2.5*\B},{1.5*\B}) {};
  \node[ecell, fill=teal!22] at ({0.5*\B},{0.5*\B}) {};
  \node[ecell, fill=teal!41] at ({1.5*\B},{0.5*\B}) {};
  \node[ecell, fill=teal!38] at ({2.5*\B},{0.5*\B}) {};
  \draw[black!80, line width=1.0pt] (0,0) rectangle ({3*\B},{3*\B});
  \foreach \k in {1,2}{\draw[gray!35, densely dashed, line width=0.35pt] ({\k*\B},0) -- ({\k*\B},{3*\B}) (0,{\k*\B}) -- ({3*\B},{\k*\B});}
  \draw[black!25, line width=0.45pt] ({\bcl-0.10},0) -- ({\bcl+2*\bs+\bw+0.10},0);
  \filldraw[fill=teal!52, draw=teal!70, line width=0.38pt] ({\bcl+0*\bs},0) rectangle ({\bcl+0*\bs+\bw},{1.000*\BH});
  \node[font=\scriptsize, anchor=north] at ({\bcl+0*\bs+0.5*\bw},{-0.14}) {$\sigma_1$};
  \filldraw[fill=teal!52, draw=teal!70, line width=0.38pt] ({\bcl+1*\bs},0) rectangle ({\bcl+1*\bs+\bw},{0.267*\BH});
  \node[font=\scriptsize, anchor=north] at ({\bcl+1*\bs+0.5*\bw},{-0.14}) {$\sigma_2$};
  \filldraw[fill=teal!52, draw=teal!70, line width=0.38pt] ({\bcl+2*\bs},0) rectangle ({\bcl+2*\bs+\bw},{0.069*\BH});
  \node[font=\scriptsize, anchor=north] at ({\bcl+2*\bs+0.5*\bw},{-0.14}) {$\sigma_3$};
  \node[font=\footnotesize\itshape, teal!75!black] at ({(\bcl+2*\bs+\bw)/2},{-0.60}) {one effective direction};
\end{scope}
\begin{scope}[xshift=7.0cm]
  \node[font=\normalsize\bfseries, anchor=south] at ({(\bcl+2*\bs+\bw)/2},{3*\B+0.42}) {High $\erk(\bm{A}) = 2.8$};
  \node[ecell, fill=teal!64] at ({0.5*\B},{2.5*\B}) {};
  \node[ecell, fill=teal!4] at ({1.5*\B},{2.5*\B}) {};
  \node[ecell, fill=teal!8] at ({2.5*\B},{2.5*\B}) {};
  \node[ecell, fill=teal!8] at ({0.5*\B},{1.5*\B}) {};
  \node[ecell, fill=teal!60] at ({1.5*\B},{1.5*\B}) {};
  \node[ecell, fill=teal!4] at ({2.5*\B},{1.5*\B}) {};
  \node[ecell, fill=teal!4] at ({0.5*\B},{0.5*\B}) {};
  \node[ecell, fill=teal!8] at ({1.5*\B},{0.5*\B}) {};
  \node[ecell, fill=teal!68] at ({2.5*\B},{0.5*\B}) {};
  \draw[black!80, line width=1.0pt] (0,0) rectangle ({3*\B},{3*\B});
  \foreach \k in {1,2}{\draw[gray!35, densely dashed, line width=0.35pt] ({\k*\B},0) -- ({\k*\B},{3*\B}) (0,{\k*\B}) -- ({3*\B},{\k*\B});}
  \draw[black!25, line width=0.45pt] ({\bcl-0.10},0) -- ({\bcl+2*\bs+\bw+0.10},0);
  \filldraw[fill=teal!52, draw=teal!70, line width=0.38pt] ({\bcl+0*\bs},0) rectangle ({\bcl+0*\bs+\bw},{1.000*\BH});
  \node[font=\scriptsize, anchor=north] at ({\bcl+0*\bs+0.5*\bw},{-0.14}) {$\sigma_1$};
  \filldraw[fill=teal!52, draw=teal!70, line width=0.38pt] ({\bcl+1*\bs},0) rectangle ({\bcl+1*\bs+\bw},{0.795*\BH});
  \node[font=\scriptsize, anchor=north] at ({\bcl+1*\bs+0.5*\bw},{-0.14}) {$\sigma_2$};
  \filldraw[fill=teal!52, draw=teal!70, line width=0.38pt] ({\bcl+2*\bs},0) rectangle ({\bcl+2*\bs+\bw},{0.739*\BH});
  \node[font=\scriptsize, anchor=north] at ({\bcl+2*\bs+0.5*\bw},{-0.14}) {$\sigma_3$};
  \node[font=\footnotesize\itshape, teal!75!black] at ({(\bcl+2*\bs+\bw)/2},{-0.60}) {three effective directions};
\end{scope}
\end{tikzpicture}}
  \end{minipage}\hfill
  \begin{minipage}[c]{0.9\columnwidth}
    \centering
    \resizebox{\linewidth}{!}{\tikzset{
  kcell/.style={minimum size=1.0cm, inner sep=1pt, align=center, font=\scriptsize, draw=teal!58, line width=0.42pt},
  kcross/.style={minimum size=1.0cm, inner sep=1pt, align=center, font=\scriptsize, draw=crimson!80, line width=0.7pt, text=crimson!85!black},
  kzero/.style={minimum size=1.0cm, inner sep=1pt, align=center, font=\scriptsize, fill=gray!7, draw=gray!25, line width=0.28pt, text=gray!40},
  gnode/.style={circle, minimum size=0.24cm, inner sep=0pt, line width=0.5pt},
}
\def\B{1.0}
\begin{tikzpicture}
\begin{scope}[xshift=0.0cm]
  \node[font=\normalsize\bfseries, anchor=south] at ({1.5*\B+1.0},{4*\B+0.2}) {Weak seam: $\integ(\bm{A}) = 0.06$};
  \node[kcell, fill=teal!60] at (0.50,3.50) {$0.80$};
  \node[kcell, fill=teal!26] at (1.50,3.50) {$0.35$};
  \node[kzero] at (2.50,3.50) {$0$};
  \node[kcell, fill=teal!45] at (0.50,2.50) {$0.60$};
  \node[kcell, fill=teal!68] at (1.50,2.50) {$0.90$};
  \node[kcross, fill=crimson!4] at (2.50,2.50) {$0.05$};
  \node[kzero] at (0.50,1.50) {$0$};
  \node[kcross, fill=crimson!4] at (1.50,1.50) {$0.05$};
  \node[kcell, fill=teal!52] at (2.50,1.50) {$0.70$};
  \node[kzero] at (0.50,0.50) {$0$};
  \node[kzero] at (1.50,0.50) {$0$};
  \node[kcell, fill=teal!38] at (2.50,0.50) {$0.50$};
  \draw[black!80, line width=1.0pt] (0,0) rectangle ({3*\B},{4*\B});
  \foreach \k in {1,2}{\draw[gray!35, densely dashed, line width=0.32pt] ({\k*\B},0) -- ({\k*\B},{4*\B});}
  \foreach \k in {1,2,3}{\draw[gray!35, densely dashed, line width=0.32pt] (0,{\k*\B}) -- ({3*\B},{\k*\B});}
  \foreach \t in {1,2,3,4} \node[font=\small, anchor=east] at ({-0.11},{(4.5-\t)*\B}) {$\bm{y}_\t$};
  \foreach \s in {1,2,3} \node[font=\small, anchor=north] at ({(\s-0.5)*\B},{-0.11}) {$\bm{x}_\s$};
  \draw[teal!45, line width=1.00pt] (3.75,3.500) -- (4.65,3.333);
  \draw[teal!22, line width=0.66pt] (3.75,3.500) -- (4.65,2.000);
  \draw[teal!35, line width=0.85pt] (3.75,2.500) -- (4.65,3.333);
  \draw[teal!50, line width=1.08pt] (3.75,2.500) -- (4.65,2.000);
  \draw[teal!40, line width=0.92pt] (3.75,1.500) -- (4.65,0.667);
  \draw[teal!30, line width=0.78pt] (3.75,0.500) -- (4.65,0.667);
  \draw[crimson!37, line width=0.44pt] (3.75,2.500) -- (4.65,0.667);
  \draw[crimson!37, line width=0.44pt] (3.75,1.500) -- (4.65,2.000);
  \draw[gray!70, densely dashed, line width=0.7pt] (3.40,2.00) -- (5.25,1.35);
  \node[font=\scriptsize\itshape, gray!60!black, anchor=west, rotate=-22] at (5.27,1.37) {seam};
  \node[gnode, fill=teal!44, draw=teal!75] at (3.75,3.500) {}; \node[font=\scriptsize, black!70, anchor=east] at (3.61,3.500) {$y_1$};
  \node[gnode, fill=teal!40, draw=teal!75] at (3.75,2.500) {}; \node[font=\scriptsize, black!70, anchor=east] at (3.61,2.500) {$y_2$};
  \node[gnode, fill=crimson!65, draw=crimson!80] at (3.75,1.500) {}; \node[font=\scriptsize, black!70, anchor=east] at (3.61,1.500) {$y_3$};
  \node[gnode, fill=crimson!70, draw=crimson!80] at (3.75,0.500) {}; \node[font=\scriptsize, black!70, anchor=east] at (3.61,0.500) {$y_4$};
  \node[gnode, fill=teal!44, draw=teal!75] at (4.65,3.333) {}; \node[font=\scriptsize, black!70, anchor=west] at (4.79,3.333) {$x_1$};
  \node[gnode, fill=teal!40, draw=teal!75] at (4.65,2.000) {}; \node[font=\scriptsize, black!70, anchor=west] at (4.79,2.000) {$x_2$};
  \node[gnode, fill=crimson!67, draw=crimson!80] at (4.65,0.667) {}; \node[font=\scriptsize, black!70, anchor=west] at (4.79,0.667) {$x_3$};
  \node[font=\footnotesize\itshape, teal!75!black] at ({1.5*\B+1.0},{-0.75}) {nearly two systems};
\end{scope}
\begin{scope}[xshift=7.0cm]
  \node[font=\normalsize\bfseries, anchor=south] at ({1.5*\B+1.0},{4*\B+0.2}) {No weak seam: $\integ(\bm{A}) = 0.65$};
  \node[kcell, fill=teal!64] at (0.50,3.50) {$0.85$};
  \node[kcell, fill=teal!30] at (1.50,3.50) {$0.40$};
  \node[kcross, fill=crimson!12] at (2.50,3.50) {$0.20$};
  \node[kcell, fill=teal!41] at (0.50,2.50) {$0.55$};
  \node[kcell, fill=teal!56] at (1.50,2.50) {$0.75$};
  \node[kcross, fill=crimson!9] at (2.50,2.50) {$0.15$};
  \node[kcell, fill=teal!22] at (0.50,1.50) {$0.30$};
  \node[kcell, fill=teal!49] at (1.50,1.50) {$0.65$};
  \node[kcross, fill=crimson!27] at (2.50,1.50) {$0.45$};
  \node[kcross, fill=crimson!42] at (0.50,0.50) {$0.70$};
  \node[kcross, fill=crimson!15] at (1.50,0.50) {$0.25$};
  \node[kcell, fill=teal!60] at (2.50,0.50) {$0.80$};
  \draw[black!80, line width=1.0pt] (0,0) rectangle ({3*\B},{4*\B});
  \foreach \k in {1,2}{\draw[gray!35, densely dashed, line width=0.32pt] ({\k*\B},0) -- ({\k*\B},{4*\B});}
  \foreach \k in {1,2,3}{\draw[gray!35, densely dashed, line width=0.32pt] (0,{\k*\B}) -- ({3*\B},{\k*\B});}
  \foreach \t in {1,2,3,4} \node[font=\small, anchor=east] at ({-0.11},{(4.5-\t)*\B}) {$\bm{y}_\t$};
  \foreach \s in {1,2,3} \node[font=\small, anchor=north] at ({(\s-0.5)*\B},{-0.11}) {$\bm{x}_\s$};
  \draw[teal!48, line width=1.04pt] (3.75,3.500) -- (4.65,3.333);
  \draw[teal!25, line width=0.70pt] (3.75,3.500) -- (4.65,2.000);
  \draw[teal!32, line width=0.81pt] (3.75,2.500) -- (4.65,3.333);
  \draw[teal!42, line width=0.96pt] (3.75,2.500) -- (4.65,2.000);
  \draw[teal!20, line width=0.62pt] (3.75,1.500) -- (4.65,3.333);
  \draw[teal!38, line width=0.89pt] (3.75,1.500) -- (4.65,2.000);
  \draw[teal!45, line width=1.00pt] (3.75,0.500) -- (4.65,0.667);
  \draw[crimson!44, line width=0.55pt] (3.75,3.500) -- (4.65,0.667);
  \draw[crimson!42, line width=0.51pt] (3.75,2.500) -- (4.65,0.667);
  \draw[crimson!55, line width=0.74pt] (3.75,1.500) -- (4.65,0.667);
  \draw[crimson!66, line width=0.92pt] (3.75,0.500) -- (4.65,3.333);
  \draw[crimson!46, line width=0.59pt] (3.75,0.500) -- (4.65,2.000);
  \draw[gray!70, densely dashed, line width=0.7pt] (3.40,1.00) -- (5.25,1.30);
  \node[font=\scriptsize\itshape, gray!60!black, anchor=west, rotate=11] at (5.27,1.32) {seam};
  \node[gnode, fill=teal!31, draw=teal!75] at (3.75,3.500) {}; \node[font=\scriptsize, black!70, anchor=east] at (3.61,3.500) {$y_1$};
  \node[gnode, fill=teal!62, draw=teal!75] at (3.75,2.500) {}; \node[font=\scriptsize, black!70, anchor=east] at (3.61,2.500) {$y_2$};
  \node[gnode, fill=gray!12, draw=gray!55] at (3.75,1.500) {}; \node[font=\scriptsize, black!70, anchor=east] at (3.61,1.500) {$y_3$};
  \node[gnode, fill=crimson!69, draw=crimson!80] at (3.75,0.500) {}; \node[font=\scriptsize, black!70, anchor=east] at (3.61,0.500) {$y_4$};
  \node[gnode, fill=gray!12, draw=gray!55] at (4.65,3.333) {}; \node[font=\scriptsize, black!70, anchor=west] at (4.79,3.333) {$x_1$};
  \node[gnode, fill=teal!58, draw=teal!75] at (4.65,2.000) {}; \node[font=\scriptsize, black!70, anchor=west] at (4.79,2.000) {$x_2$};
  \node[gnode, fill=crimson!70, draw=crimson!80] at (4.65,0.667) {}; \node[font=\scriptsize, black!70, anchor=west] at (4.79,0.667) {$x_3$};
  \node[font=\footnotesize\itshape, teal!75!black] at ({1.5*\B+1.0},{-0.75}) {hard to split};
\end{scope}
\end{tikzpicture}}
  \end{minipage}
  \caption{%
    \textbf{Top:} Effective rank $\erk(\bm{A})$.
    A rank-1-like matrix has one dominant singular value and $\erk=1.2$;
    a full-rank matrix has comparable singular values and $\erk=2.8$.
    \textbf{Bottom:} Integrity $\integ(\bm{A})$, the weakest seam.
    Nodes of the graph $G(\bm{A})$ are coloured by the Fiedler vector: teal and burgundy mark the two halves the system most nearly splits into, pale nodes sit on the seam (dashed).
    Burgundy entries and edges cross the seam.
    Left: two weakly connected blocks.
    Right: in a dense matrix integrity is higher.}
  \label{fig:erk_kc}
  \figalttext[Effective rank and integrity on small example matrices.]{Top: two matrices with their singular values; a nearly rank-one matrix has effective rank close to 1 and a full-rank matrix has effective rank about 2.8. Bottom: bipartite graphs of two matrices with nodes coloured by the Fiedler vector, showing the two halves along which each system most nearly splits; a weak seam gives low integrity.}%
\end{figure}

\paragraph{Scale and sensitivity.}
Uniformly attenuating a nonzero Jacobian leaves both measures unchanged, even as its magnitude approaches zero. They measure relative structure, not absolute sensitivity. Fading differences therefore require selective attenuation or an explicit noise model, as in the fog example, \secref{ssec:boundaries}.

\paragraph{The constitution of boundaries.}
Hypothesis~\ref{h:first} grades experience by transparency and integrity, but does not say where one experiencing system ends and another begins: whose experience it is. A possible approach, which requires empirical validation, is to treat units, defined as follows, as candidates for subjects.
 
 \begin{definition}
  \label{def:boundary}
  A \emph{candidate unit} $X$ is a set of output nodes (spacetime coordinates), parameters, and input nodes through which every path into the outputs passes. Two candidates are \emph{neighbours} if one is obtained from the other by adding/removing an output node, temporal coordinate or parameter, by replacing input nodes with upstream nodes, or replacing output nodes with downstream nodes. A \textbf{unit} is a candidate that locally maximises integral rank,
  \begin{equation}
    \label{eq:unit}
    \integ_{\erk}(X)\;\ge\;\integ_{\erk}(X')\quad\text{for all neighbours } X',
  \end{equation}
  where adding/removing output nodes and parameters is scored on $\integ_{\erk}(\bJ_{\by\leftarrow\cW})$, and changing temporal coordinates or replacing input or output nodes on $\integ_{\erk}(\bJ_{\by\leftarrow\bx})$.
\end{definition}

\noindent
Parameters are stable over the timeframe, \secref{ssec:ops}; every temporal coordinate shares them. Changes to temporal coordinates are therefore scored on the input Jacobian. The input cut may reach further into the past than the outputs.

The examples below apply \eqref{eq:unit} to three coarse moves:
\begin{enumerate}
  \item \emph{merge (\S\ref{ssec:world_in_me}):}
  $A\cup B$ is one unit iff
  $\integ_{\erk}(A\cup B) \ge \max\{\integ_{\erk}(A),\integ_{\erk}(B)\}$;
  \item \emph{extend (\S\ref{ssec:boundaries}):}
  the input cut moves upstream through a segment $\bS$ iff
  $\integ_{\erk}(\bJ\bS)\ge\integ_{\erk}(\bJ)$, and the output cut moves downstream through $\bS$ iff $\integ_{\erk}(\bS\bJ)\ge\integ_{\erk}(\bJ)$;
  \item \emph{duration (\S\ref{ssec:time}):}
  for outputs $\by_{t-T+1:t}$, with the cut at the whole past that makes a difference, the present is the $T$ that locally maximises $\integ_{\erk}\big(\bJ_{\by_{t-T+1:t}\leftarrow\bx_{\le t}}\big)$ and beats the baseline of a single moment, below.
\end{enumerate}
Ties are indeterminate: a transmissive segment (one that relabels and uniformly rescales, \secref{ssec:boundaries}) gives equality, and the boundary may sit at either end. Integrity alone favours tiny candidates; transparency alone favours large ones that need not interact. A candidate with a single output, or a single input, has a rank-one Jacobian and integral rank exactly $1$: perfectly integrated but registering one difference. It is a unit whenever its neighbours score below $1$. An \emph{extended unit} is a unit that beats this baseline: a candidate subject whose experience extends beyond a single difference.

\section{Results}
\label{sec:life}

The examples below are neural nets, for which standard libraries such as PyTorch \citep{paszke2019} compute Jacobians exactly. They concern units, \secref{ssec:world_in_me}--\ref{ssec:boundaries}; extent in time and space, \secref{ssec:time}--\ref{ssec:space}; structure within experience, \secref{ssec:facets}--\ref{ssec:distinct}; and learning and hardware, \secref{ssec:learn}--\ref{ssec:hardware}. \secref{ssec:world_in_me}--\ref{ssec:space} use transparency, integrity and integral rank directly; the rest look inside the Jacobian, at the rows, entries and covariances that the two measures summarise.

\subsection{Shared inputs; independent decisions}
\label{ssec:world_in_me}

\begin{figure}[ht]
  \centering
  \resizebox{0.95\columnwidth}{!}{\tikzset{
  tcanopy/.style={circle, fill=teal!20, draw=teal!55, line width=0.5pt},
  ttrunk/.style={fill=teal!42, draw=teal!60, line width=0.4pt},
  thid/.style={circle, minimum size=0.28cm, inner sep=0pt,
    fill=teal!42, draw=teal!62, line width=0.45pt},
  tout/.style={circle, minimum size=0.28cm, inner sep=0pt,
    fill=teal!42, draw=teal!62, line width=0.45pt},
  twnode/.style={circle, minimum size=0.3cm, inner sep=0pt,
    fill=teal!85!black, draw=teal!85!black, line width=0.9pt},
  aedge/.style={teal!42, line width=1.0pt},
  bedge/.style={teal!85!black, line width=1.1pt, dash pattern=on 2.6pt off 1.9pt},
  tcell/.style={minimum size=1.0cm, inner sep=1pt, align=center, font=\scriptsize,
    draw=teal!58, line width=0.42pt},
  tzero/.style={minimum size=1.0cm, inner sep=1pt, align=center, font=\scriptsize,
    fill=gray!7, draw=gray!25, line width=0.28pt, text=gray!40},
}

\newcommand{\pertburst}[3]{%
  \foreach \ang in {0,45,90,135,180,225,270,315}{
    \draw[#3, line width=0.55pt] (#1,#2) -- ++(\ang:0.16);
  }
  \filldraw[#3] (#1,#2) circle (0.05);
}

\begin{tikzpicture}

\begin{scope}[yshift=5.3cm]

  \node[tcanopy, minimum size=0.9cm] (canopy) at (5.7,4.1) {};
  \draw[ttrunk] (5.6,3.5) rectangle (5.8,3.8);

  \node[thid] (yh1) at (2.0,2.6) {};
  \node[thid] (yh2) at (2.0,1.9) {};
  \node[tout] (y1a) at (1.65,1.0) {};
  \node[tout] (y1b) at (2.35,1.0) {};

  \node[thid] (mh1) at (9.4,2.6) {};
  \node[thid] (mh2) at (9.4,1.9) {};
  \node[tout] (y2a) at (9.05,1.0) {};
  \node[tout] (y2b) at (9.75,1.0) {};

  \foreach \tp in {5.65,5.75}{
    \draw[aedge] (\tp,3.5) -- (yh1);
    \draw[aedge] (\tp,3.5) -- (yh2);
    \draw[aedge] (\tp,3.5) -- (mh1);
    \draw[aedge] (\tp,3.5) -- (mh2);
  }
  \draw[aedge] (yh1) -- (y1a);
  \draw[aedge] (yh1) -- (y1b);
  \draw[aedge] (yh2) -- (y1a);
  \draw[aedge] (yh2) -- (y1b);
  \draw[aedge] (mh1) -- (y2a);
  \draw[aedge] (mh1) -- (y2b);
  \draw[aedge] (mh2) -- (y2a);
  \draw[aedge] (mh2) -- (y2b);

  \pertburst{5.7}{4.6}{teal!70!black}
  \node[font=\scriptsize, teal!60!black, anchor=west] at (6.25,4.6) {perturb $\bm{x}$: the tree};

  \draw[bedge] (yh1) -- (y1a);
  \node[twnode] at (yh1) {};
  \node[tout, draw=teal!85!black, line width=0.8pt] at (y1a) {};
  \pertburst{1.82}{1.8}{teal!85!black}
  \node[font=\scriptsize, teal!85!black, anchor=east] at (1.55,2.05) {perturb $v_{jk}$};

  \node[font=\small, anchor=north] at (2.0,{0.7}) {you: $\bm{y}_1$};
  \node[font=\small, anchor=north] at (9.4,{0.7}) {me: $\bm{y}_2$};

\end{scope}

\begin{scope}

  \node[font=\footnotesize, anchor=south, teal!42!black] at (2,4.2) {input Jacobian: perturb $\bm{x}$};

  \node[tcell, fill=teal!58] at (0.5,3.5) {$0.82$};
  \node[tcell, fill=teal!24] at (1.5,3.5) {$0.32$};
  \node[tcell, fill=teal!41] at (2.5,3.5) {$0.55$};
  \node[tcell, fill=teal!15] at (3.5,3.5) {$0.20$};

  \node[tcell, fill=teal!36] at (0.5,2.5) {$0.48$};
  \node[tcell, fill=teal!63] at (1.5,2.5) {$0.86$};
  \node[tcell, fill=teal!20] at (2.5,2.5) {$0.27$};
  \node[tcell, fill=teal!44] at (3.5,2.5) {$0.60$};

  \node[tcell, fill=teal!30] at (0.5,1.5) {$0.40$};
  \node[tcell, fill=teal!52] at (1.5,1.5) {$0.72$};
  \node[tcell, fill=teal!66] at (2.5,1.5) {$0.90$};
  \node[tcell, fill=teal!18] at (3.5,1.5) {$0.24$};

  \node[tcell, fill=teal!22] at (0.5,0.5) {$0.30$};
  \node[tcell, fill=teal!47] at (1.5,0.5) {$0.65$};
  \node[tcell, fill=teal!38] at (2.5,0.5) {$0.52$};
  \node[tcell, fill=teal!70] at (3.5,0.5) {$0.95$};

  \draw[black!80, line width=1.0pt] (0,0) rectangle (4,4);
  \foreach \k in {1,2,3}{
    \draw[gray!35, densely dashed, line width=0.32pt]
      (\k,0) -- (\k,4)
      (0,\k) -- (4,\k);
  }
  \node[font=\small, anchor=east] at (-0.12,3.5) {$y_{1a}$};
  \node[font=\small, anchor=east] at (-0.12,2.5) {$y_{1b}$};
  \node[font=\small, anchor=east] at (-0.12,1.5) {$y_{2a}$};
  \node[font=\small, anchor=east] at (-0.12,0.5) {$y_{2b}$};
  \foreach \s/\lab in {1/x_1, 2/x_2, 3/x_3, 4/x_4}
    \node[font=\small, anchor=north] at ({\s-0.5},-0.12) {$\lab$};

  \draw[decorate, decoration={brace, amplitude=4pt}]
    (-0.75,2) -- (-0.75,4) node[midway, left=6pt, font=\scriptsize, rotate=90, anchor=south] {you};
  \draw[decorate, decoration={brace, amplitude=4pt}]
    (-0.75,0) -- (-0.75,2) node[midway, left=6pt, font=\scriptsize, rotate=90, anchor=south] {me};

  \node[font=\footnotesize\itshape, teal!75!black] at (2,-0.7)
    {$\bm{J}_{\{\bm{y}_1,\bm{y}_2\}\leftarrow\bm{x}} $: $\integ>0$, tangled};

\end{scope}

\begin{scope}[xshift=7.4cm]

  \node[font=\footnotesize, anchor=south, teal!85!black] at (2,4.2) {parameter Jacobian: perturb $v_{jk}$ or $w_{jk}$};

  \node[tcell, fill=teal!60] at (0.5,3.5) {$0.80$};
  \node[tcell, fill=teal!28] at (1.5,3.5) {$0.35$};
  \node[tzero] at (2.5,3.5) {$0$};
  \node[tzero] at (3.5,3.5) {$0$};

  \node[tcell, fill=teal!45] at (0.5,2.5) {$0.60$};
  \node[tcell, fill=teal!68] at (1.5,2.5) {$0.90$};
  \node[tzero] at (2.5,2.5) {$0$};
  \node[tzero] at (3.5,2.5) {$0$};

  \node[tzero] at (0.5,1.5) {$0$};
  \node[tzero] at (1.5,1.5) {$0$};
  \node[tcell, fill=teal!53] at (2.5,1.5) {$0.70$};
  \node[tcell, fill=teal!24] at (3.5,1.5) {$0.30$};

  \node[tzero] at (0.5,0.5) {$0$};
  \node[tzero] at (1.5,0.5) {$0$};
  \node[tcell, fill=teal!33] at (2.5,0.5) {$0.42$};
  \node[tcell, fill=teal!58] at (3.5,0.5) {$0.78$};

  \draw[teal!70, line width=0.95pt] (0,2) rectangle (2,4);
  \draw[teal!70, line width=0.95pt] (2,0) rectangle (4,2);

  \draw[black!80, line width=1.0pt] (0,0) rectangle (4,4);
  \foreach \k in {1,2,3}{
    \draw[gray!35, densely dashed, line width=0.32pt]
      (\k,0) -- (\k,4)
      (0,\k) -- (4,\k);
  }
  \node[font=\small, anchor=east] at (-0.12,3.5) {$y_{1a}$};
  \node[font=\small, anchor=east] at (-0.12,2.5) {$y_{1b}$};
  \node[font=\small, anchor=east] at (-0.12,1.5) {$y_{2a}$};
  \node[font=\small, anchor=east] at (-0.12,0.5) {$y_{2b}$};
  \foreach \s/\lab in {1/v_1, 2/v_2, 3/w_1, 4/w_2}
    \node[font=\small, anchor=north] at ({\s-0.5},-0.12) {$\lab$};

  \draw[decorate, decoration={brace, amplitude=4pt}]
    (-0.75,2) -- (-0.75,4) node[midway, left=6pt, font=\scriptsize, rotate=90, anchor=south] {you};
  \draw[decorate, decoration={brace, amplitude=4pt}]
    (-0.75,0) -- (-0.75,2) node[midway, left=6pt, font=\scriptsize, rotate=90, anchor=south] {me};

  \node[font=\footnotesize\itshape, teal!75!black] at (2,-0.7)
    {$\bm{J}_{\{\bm{y}_1,\bm{y}_2\}\leftarrow\{\cV,\cW\}}$: $\integ=0$, independent};

\end{scope}

\end{tikzpicture}}
  \caption{Two MLPs see a tree.
    \textbf{Top:} perturbing the input propagates into both nets; perturbing a parameter stays confined to one net.
    \textbf{Bottom left:} the input Jacobian $\bJ_{\{\by_1,\by_2\}\leftarrow \bx}$ is dense: the two nets are tangled.
    \textbf{Bottom right:} the parameter Jacobian $\bJ_{\{\by_1,\by_2\}\leftarrow \{\cV,\cW\}}$ is block diagonal: the two nets' decisions are independent.}
  \label{fig:tree_jacs}
  \figalttext[Two networks looking at the same tree, with their input and parameter Jacobians.]{Top: a perturbation of the shared input spreads into both networks, while a perturbation of one network's parameters stays inside that network. Bottom left: the joint input Jacobian is dense. Bottom right: the joint parameter Jacobian is block diagonal, one block per network.}%
\end{figure}

\paragraph{You and me, \emph{we} see a tree.}
To understand the distinct roles of the input and parameter Jacobians, consider two MLPs looking at the same image, Fig.~\ref{fig:tree_jacs}. If you perturb the image, the outputs of both MLPs change. Their input Jacobians are interwoven:
\begin{equation*}
  \integ\big(\bJ_{\{\by_1,\by_2\}\leftarrow \bx}\big) > 0
\end{equation*}
However, perturbing the parameters of one MLP does not affect the outputs of the other, so their parameter Jacobians are independent:
\begin{equation*}
  \integ\big(\bJ_{\{\by_1,\by_2\}\leftarrow \{\cV,\cW\}}\big) = 0.
\end{equation*}
If you and I look at the same side of the same tree, what is going on in your head has something, physically, to do with what is going on in mine; on Hypothesis~\ref{h:first}, \textbf{our experiences overlap} \citep{james1905}. The mathematics captures that: the input Jacobian from the tree to the outputs of both nets has nonzero integrity.

Yet \textbf{you are not me}. Your decisions are independent of mine (we have never met, we have no shared past). The mathematics captures that: the parameter Jacobian of the nets decomposes into independent sub-Jacobians.

\paragraph{I see a tree.}
When I look at a tree, I would like to say that I see the tree itself, not a pattern on my retina or a representation in my visual cortex \citep{skokowski2022}. There are lines of force from the tree into my brain: perturb the tree and my brain registers it. If I see the tree, then I, as an experiencing physical system, must encompass parts of my brain \emph{and} parts of the tree. Seeing through retina and cortex is learned: after childhood blindness, the newly sighted adapt with difficulty at best \citep{gregory1969}.

\paragraph{Interacting systems.}
If you and I interact, then my outputs become your inputs, and changing my parameters changes, slightly, what you do. Our joint parameter Jacobian is connected, with a weak seam between us. A conversation should not make us one person; under Definition~\ref{def:boundary} it does not: units merge only when the coupling between them is comparable to the coupling within. Take two small MLPs trained separately on handwritten digits, with each one's hidden layer feeding the other's at strength $\epsilon$. Over ten pairs of nets, at ten inputs each, integral rank keeps them apart at $\epsilon\le1$ in every case and merges them at $\epsilon=2$ in $96$ of $100$; if only one feeds the other they stay apart at every strength tried, \secref{app:integrity}.

\subsection{Focus can select boundaries}
\label{ssec:boundaries}

\secref{ssec:world_in_me} claimed that seeing a tree involves a system encompassing parts of the brain \emph{and} parts of the tree. This section analyses boundaries in a simple model of the eye.

Suppose I look at a tree. For any point $c$ on the path --- retina, cornea, air --- the chain rule gives $\bJ_{\by\leftarrow\textrm{tree}} = \bJ_{\by\leftarrow c}\cdot\bJ_{c\leftarrow\textrm{tree}}$. Call the Jacobian between two such points a \textbf{segment}. A segment that is a positive multiple of a permutation is \textbf{transmissive}: it relabels and uniformly rescales without mixing --- the limit case of \emph{seeing through}. Both measures are scale and permutation invariant, \secref{ssec:analysis}, so including a transmissive segment changes nothing and the boundary can be placed at either end. \textbf{Systems are individuated by where they start mixing.} By symmetry, the outputs extend through a transmissive segment downstream: a system acts through a limb, or a well-held tool, as it sees through a focused eye.

\begin{figure}[t]
  \centering
  \includegraphics[width=\columnwidth]{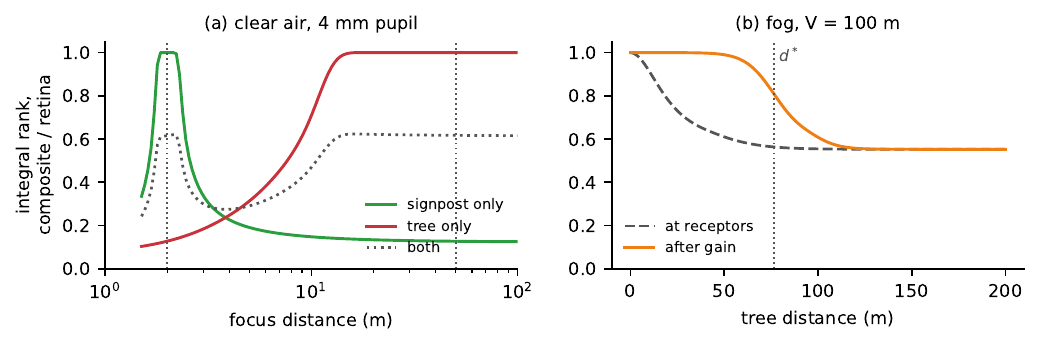}
  \caption{\textbf{Focus can select boundaries.} Integral rank of the composite system, downstream Jacobian $\times$ segment, relative to the downstream Jacobian alone (boundary at the retina), \S\ref{app:optics}. \textbf{(a)} Clear air. With a single object as input, its segment costs nothing when it is in focus and most of the integral rank when it is not: focus selects which boundary is admissible. With both objects as input, no focus frees both segments. \textbf{(b)} Fog, $100\,$m visibility; the tree, with the in-focus signpost alongside. At the photoreceptors the segment's cost grows with distance. After optimal contrast restoration it is negligible until noise wins at $d^*$.}
  \label{fig:boundaries}
  \figalttext[Bar charts of integral rank when the boundary is moved through in-focus and out-of-focus segments, in clear air and in fog.]{Integral rank of the composite system relative to the boundary at the retina, for single objects and for two objects with the eye focused on one. In-focus segments cost nothing, out-of-focus segments cost most of the integral rank, and in fog the boundary is pulled in.}%
\end{figure}

When out of focus, each photoreceptor sums light from several points of an object and the segment mixes; when in focus, the light from each point lands on one photoreceptor and the segment is transmissive. The eye focuses at one depth at a time, so it chooses which segment transmits. Definition~\ref{def:boundary} makes this precise. Including an in-focus object's segment leaves the composite system's integral rank exactly unchanged. With a single object as input, including its segment out of focus lowers integral rank by $87\%$ on average, Fig.~\ref{fig:boundaries}a. With both objects in view and the eye focused on a nearby signpost, the cut moves through the signpost's segment at no cost, while moving it through the distant tree's segment costs $38\%$ on average, and lowers it in every draw: the boundary can extend to the signpost, while the tree is cut off at the retina. Refocus and the two swap. \textbf{The input boundary is actively selected by focusing.} Selection works object by object: no focus makes both segments free.

Fog scatters light, mixing transmission through the air and weakening the unscattered signal relative to the receptors' noise. Downstream contrast restoration can undo most of the mixing, but amplifies that noise. After optimal restoration, including the tree's segment is essentially free out to half the visibility distance and costly beyond $d^*\approx0.77$ of it, where noise wins: \textbf{fog pulls the input boundary in}, Fig.~\ref{fig:boundaries}b.

\subsection{The structure of duration}
\label{ssec:time}

A clock chimes five times. The sounds are identical; the experience of hearing them is not. The experience of the second chime is coloured by the first, the third by the first two and so on. Time is not experienced in addition to the five chimes; the experience of time is the experience of their succession. Integrity, and with it integral rank, makes it possible to understand how interactions weave the past into the present.

Compare the input Jacobians $\bJ_{\by_{1:T}\leftarrow \bx_{1:T}}$ of an MLP and an RNN over a time series, Fig.~\ref{fig:rnn_jacs}. For simplicity suppose both MLP and RNN have one-dimensional input and output, and higher-dimensional hidden layers. The Jacobians are $T\times T$ matrices where columns correspond to perturbed inputs at times $1,\ldots, T$ and rows correspond to outputs.

The MLP's Jacobian is diagonal because there are no interactions across time; with at least two active instants its integrity is zero. In the illustrated RNN, interactions across time connect the active graph, giving positive integrity: the influence of the past endures into the present. The recurrent net unfolds as a feedforward (directed acyclic) graph over spacetime, as in the unfolding argument in \citet{doerig2019}. Duration arises from how inputs at different times are mixed.

\begin{figure}[htbp]
  \centering
  \resizebox{0.9\columnwidth}{!}{\tikzset{
    diag/.style={
      minimum size=1.45cm, inner sep=2pt, align=center, font=\small,
      fill=teal!32, draw=teal!65, line width=0.55pt
    },
    rec/.style={
      minimum size=1.45cm, inner sep=2pt, align=center, font=\small,
      fill=teal!14, draw=teal!50, line width=0.4pt
    },
    zero/.style={
      minimum size=1.45cm, inner sep=2pt, align=center, font=\small,
      fill=gray!6, draw=gray!28, line width=0.3pt, text=gray!45
    },
    inode/.style={circle, minimum size=0.24cm, inner sep=0pt,
      fill=teal!20, draw=teal!55, line width=0.42pt},
    hnode/.style={circle, minimum size=0.24cm, inner sep=0pt,
      fill=teal!48, draw=teal!72, line width=0.42pt},
    snode/.style={circle, minimum size=0.32cm, inner sep=0pt,
      fill=teal!50, draw=teal!75, line width=0.55pt},
    onode/.style={circle, minimum size=0.24cm, inner sep=0pt,
      fill=teal!20, draw=teal!55, line width=0.42pt},
    fedge/.style={teal!55, line width=0.38pt},
    recurse/.style={-{Stealth[length=3.5pt, width=2.8pt]},
      teal!68, line width=0.62pt},
  }
  \def\B{1.45}
  \def\snr{0.16}
  \begin{tikzpicture}

  \pgfmathsetmacro{\ynet}{4*\B + 0.45}
  \pgfmathsetmacro{\yin}{\ynet + 0.28}
  \pgfmathsetmacro{\yhid}{\ynet + 1.10}
  \pgfmathsetmacro{\yout}{\ynet + 1.92}
  \pgfmathsetmacro{\ytitle}{\ynet + 2.80}

  \begin{scope}

    \foreach \s in {1,...,4}{
      \pgfmathsetmacro{\cx}{(\s-0.5)*\B}
      \draw[fedge] (\cx, \yin)         -- ({\cx-0.27}, \yhid);
      \draw[fedge] (\cx, \yin)         -- ({\cx+0.27}, \yhid);
      \draw[fedge] ({\cx-0.27}, \yhid) -- (\cx, \yout);
      \draw[fedge] ({\cx+0.27}, \yhid) -- (\cx, \yout);
    }
    \foreach \s in {1,...,4}{
      \pgfmathsetmacro{\cx}{(\s-0.5)*\B}
      \node[inode] at (\cx, \yin)         {};
      \node[hnode] at ({\cx-0.27}, \yhid) {};
      \node[hnode] at ({\cx+0.27}, \yhid) {};
      \node[onode] at (\cx, \yout)        {};
    }
    \node[font=\scriptsize, teal!60!black, anchor=north]
      at ({0.5*\B}, {\yin - 0.19}) {$\bm{x}_t$};
    \node[font=\scriptsize, teal!60!black, anchor=south]
      at ({0.5*\B}, {\yout + 0.19}) {$\bm{y}_t$};

    \node[font=\normalsize\bfseries, anchor=south] at ({2*\B},\ytitle)
      {MLP\enspace
       $\bm{J}_{\bm{y}_{1:T}\leftarrow\bm{x}_{1:T} } = \nabla_{\bm{x}_{1:T}}(\bm{y}_{1:T})$};

    \foreach \t in {1,...,4}{
      \foreach \s in {1,...,4}{
        \pgfmathparse{int(\t==\s)}
        \ifnum\pgfmathresult=1
          \node[diag] at ({(\s-0.5)*\B},{(4.5-\t)*\B})
            {$\dfrac{\partial\bm{y}_\t}{\partial\bm{x}_\t}$};
        \else
          \node[zero] at ({(\s-0.5)*\B},{(4.5-\t)*\B})
            {$\bm{0}$};
        \fi
      }
    }

    \draw[black!80, line width=1.1pt] (0,0) rectangle ({4*\B},{4*\B});
    \foreach \k in {1,2,3}{
      \draw[gray!35, densely dashed, line width=0.35pt]
        ({\k*\B},0)     -- ({\k*\B},{4*\B})
        (0,{\k*\B})     -- ({4*\B},{\k*\B});
    }
    \foreach \t in {1,...,4}
      \node[font=\small, anchor=east] at ({-0.12},{(4.5-\t)*\B})
        {$\bm{y}_\t$};
    \foreach \s in {1,...,4}
      \node[font=\small, anchor=north] at ({(\s-0.5)*\B},{-0.12})
        {$\bm{x}_\s$};
    \draw[decorate, decoration={brace, amplitude=5pt, mirror}]
      (0,{-0.65}) -- ({4*\B},{-0.65})
      node[midway, below=7pt, font=\footnotesize] {$dT$};
    \draw[decorate, decoration={brace, amplitude=5pt}]
      ({-0.8},0) -- ({-0.8},{4*\B})
      node[midway, left=8pt, font=\footnotesize, rotate=90, anchor=south] {$nT$};

  \end{scope}

  \begin{scope}[xshift=8.4cm]

    \foreach \s in {1,...,4}{
      \pgfmathsetmacro{\cx}{(\s-0.5)*\B}
      \draw[fedge] (\cx, \yin)  -- (\cx, \yhid);
      \draw[fedge] (\cx, \yhid) -- (\cx, \yout);
    }
    \foreach \s in {1,...,4}{
      \pgfmathsetmacro{\cx}{(\s-0.5)*\B}
      \node[inode] at (\cx, \yin)  {};
      \node[snode] at (\cx, \yhid) {};
      \node[onode] at (\cx, \yout) {};
    }
    \foreach \s [evaluate=\s as \snext using int(\s+1)] in {1,2,3}{
      \pgfmathsetmacro{\xstart}{(\s     - 0.5)*\B + \snr}
      \pgfmathsetmacro{\xend}  {(\snext - 0.5)*\B - \snr}
      \draw[recurse] (\xstart, \yhid) -- (\xend, \yhid);
    }
    \node[font=\scriptsize, teal!60!black, anchor=north]
      at ({0.5*\B}, {\yin - 0.19}) {$\bm{x}_t$};
    \node[font=\scriptsize, teal!60!black, anchor=south]
      at ({0.5*\B}, {\yout + 0.19}) {$\bm{y}_t$};
    \node[font=\scriptsize, teal!60!black, anchor=south]
      at ({0.7*\B}, {\yhid + 0.12}) {$\bm{h}_t$};

    \node[font=\normalsize\bfseries, anchor=south] at ({2*\B},\ytitle)
      {RNN\enspace
       $\bm{J}_{\bm{y}_{1:T}\leftarrow \bm{x}_{1:T}} = \nabla_{\bm{x}_{1:T}}(\bm{y}_{1:T})$};

    \foreach \t in {1,...,4}{
      \foreach \s in {1,...,4}{
        \pgfmathparse{int(\t>=\s)}
        \ifnum\pgfmathresult=1
          \pgfmathparse{int(\t==\s)}
          \ifnum\pgfmathresult=1
            \node[diag] at ({(\s-0.5)*\B},{(4.5-\t)*\B})
              {$\dfrac{\partial\bm{y}_\t}{\partial\bm{x}_\t}$};
          \else
            \node[rec] at ({(\s-0.5)*\B},{(4.5-\t)*\B})
              {$\dfrac{\partial\bm{y}_\t}{\partial\bm{x}_\s}$};
          \fi
        \else
          \node[zero] at ({(\s-0.5)*\B},{(4.5-\t)*\B})
            {$\bm{0}$};
        \fi
      }
    }

    \draw[black!80, line width=1.1pt] (0,0) rectangle ({4*\B},{4*\B});
    \foreach \k in {1,2,3}{
      \draw[gray!35, densely dashed, line width=0.35pt]
        ({\k*\B},0)     -- ({\k*\B},{4*\B})
        (0,{\k*\B})     -- ({4*\B},{\k*\B});
    }
    \foreach \t in {1,...,4}
      \node[font=\small, anchor=east] at ({-0.12},{(4.5-\t)*\B})
        {$\bm{y}_\t$};
    \foreach \s in {1,...,4}
      \node[font=\small, anchor=north] at ({(\s-0.5)*\B},{-0.12})
        {$\bm{x}_\s$};
    \draw[decorate, decoration={brace, amplitude=5pt, mirror}]
      (0,{-0.65}) -- ({4*\B},{-0.65})
      node[midway, below=7pt, font=\footnotesize] {$dT$};

  \end{scope}
  \end{tikzpicture}}
  \caption{Jacobians for MLP and RNN over length $T$ sequence.
    \textbf{MLP} (left): each output $\bm{y}_t$ depends only on its own input $\bm{x}_t$, giving a diagonal Jacobian.
    \textbf{RNN} (right): hidden state carries information across time, giving a lower-triangular Jacobian where $\bm{y}_t$ depends on $\bm{x}_{1:t}$.}
  \label{fig:rnn_jacs}
  \figalttext[Input Jacobians over a sequence for an MLP and an RNN.]{Left: the MLP's Jacobian over a length-T sequence is diagonal, since each output depends only on its own input. Right: the RNN's Jacobian is lower triangular, since each output depends on all earlier inputs.}%
\end{figure}

\paragraph{Quantifying duration.}
How the past endures depends on the size of the state and the update, Eq.~\eqref{eq:rnn}. Consider a minimal example, where channels compute exponential moving averages (EMAs):
\begin{equation}
  \label{eq:ema}
  h^{(i)}_t = \delta_i\cdot h^{(i)}_{t-1} + (1-\delta_i)\cdot x_t,
  \qquad
  y_t = \textstyle\sum_i h^{(i)}_t,
\end{equation}
with fixed decays $\delta_i\in(0,1)$, one per channel.

\begin{figure}[t]
  \centering
  \includegraphics[width=0.98\columnwidth]{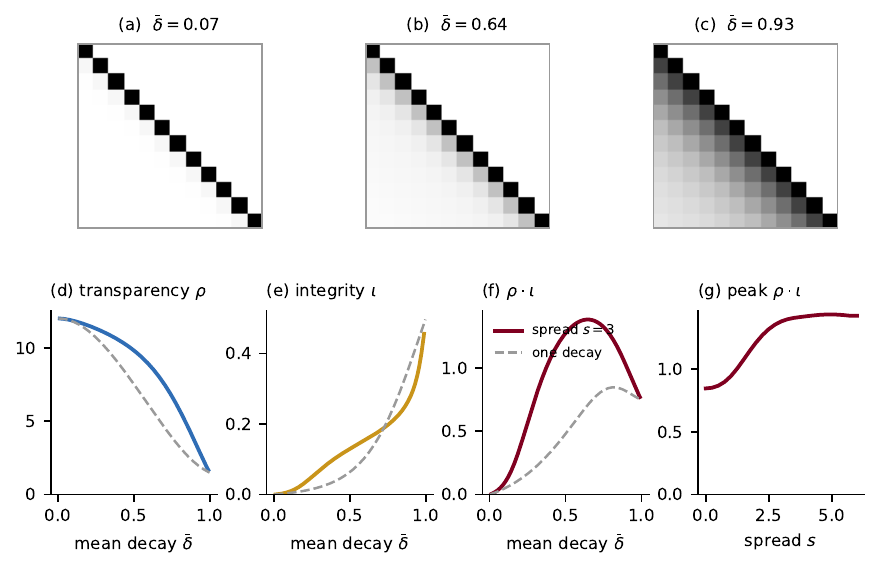}
  \caption{\textbf{Quantifying duration.}
  Multi-timescale EMA, Eq.~\eqref{eq:ema}, six channels on a logistic ladder with spread $s=3$, window $T=12$.
  \textbf{Top:} $\lvert\bJ_{y_{1:T}\leftarrow x_{1:T}}\rvert$ rescaled to unit mean at three points of the sweep. \textbf{(a)} $\bar\delta=0.07$: all channels fast, near-diagonal. \textbf{(b)} $\bar\delta=0.64$: timescales spread across the window, a graded triangle. \textbf{(c)} $\bar\delta=0.93$: all channels slow, near-uniform below the diagonal.
  \textbf{Bottom:} sweeping the mean decay $\bar\delta$. \textbf{(d)} Transparency falls as the past smears into the present. \textbf{(e)} Integrity rises as the past is tied to the present. \textbf{(f)} Their product, integral rank, peaks in the interior. Dashed grey: a single decay. \textbf{(g)} Peak integral rank against the spread of timescales $s$.}
  \label{fig:ema_duration}
  \figalttext[Jacobians and integral rank for multi-timescale exponential moving averages.]{Top: absolute Jacobians of a six-channel EMA at three mean decays, from near-diagonal to a graded triangle to a near-uniform lower triangle when all channels are slow. Bottom: transparency, integrity and integral rank across the decay sweep, and the effect of spreading the decays.}%
\end{figure}

Place the decays on a logistic ladder, $\delta_i=\mathrm{logistic}(\mu+s\,z_i)$, with $z_1,\ldots,z_6$ evenly spaced on $[-1,1]$: $\mu$ sets the mean decay and $s$ the spread of timescales, with $s=0$ a single decay. Fig.~\ref{fig:ema_duration} sweeps $\mu$ over a window of $T=12$ steps. A slow decay is temporal blur, as defocus is spatial blur, \secref{ssec:boundaries}. The difference is that nothing else mixes across time: without interaction across time the input Jacobian is diagonal and its integrity is zero. Temporal blur is the only source of integrity. Transparency falls monotonically: the more the past smears into the present, the fewer independent moments get through. Integrity rises monotonically. Neither extreme maximises integral rank, for the reason in \secref{ssec:analysis}. The near-diagonal Jacobian is transparent but falls apart into instants. The dense triangle is hard to split but opaque: its rows are nested sums that pile up along one direction.

\textbf{Spreading the decays improves the trade-off}, Fig.~\ref{fig:ema_duration}f--g. At the same mean decay, channels at different speeds keep more of the past distinct, because contributions at different timescales do not lie in one another's span. Over the fixed window $T=12$, peak integral rank rises from $0.85$ with a single decay to about $1.4$ with a wide spread. 

\paragraph{EMAs select the experienced present.}
The decay sweep compares dynamics; Definition~\ref{def:boundary} selects the present. With a single timescale, integral rank over windows of several steps stays below the baseline of $1$ set by a single moment (\S\ref{ssec:analysis}): \textbf{a single timescale selects an instantaneous present}. Spreading the timescales changes this. Six channels with spread $s=3$ select a present of about three times their slowest timescale $-1/\log\delta_{\max}$. With slowest timescale $50$, the selected present is $155$ steps, well above the baseline: an extended unit. \textbf{An extended present may need a spread of timescales.} Brains are massively recurrent, and in cortex synaptic, cellular and network time constants vary tremendously \citep{koch1996,magee2000,spruston2008}, with intrinsic timescales and temporal receptive windows lengthening up the cortical hierarchy \citep{murray2014,hasson2008}.

\subsection{Spatial structure: convolution}
\label{ssec:space}

A simple model of spatial perception is a 2D-convolution:
\begin{equation*}
  z_{x,y} = \bk * h_{x,y}
  = \sum_{i=-a}^a\sum_{j=-b}^b k_{ij} \cdot h_{x-i,y-j}
\end{equation*}
where $\bk$ is a $(2a+1)\times (2b+1)$ kernel.
The input Jacobian is the 4-tensor
\begin{equation*}
  \bJ_{(x,y)\leftarrow(p,q)}
  = k_{x-p, y-q}
\end{equation*}
In other words, for each output coordinate $(x,y)$ the Jacobian recovers the kernel, flipped and centred at $(x,y)$ in $\bh$-space. The kernel is \emph{how} the grid is mixed: how neighbouring points are interwoven.

\paragraph{Closed form expressions.}
On an $N\times N$ grid with periodic boundary, both measures have closed forms, \S\ref{app:conv}. Let $\hat k(\omega)$ be the Fourier transform of $\bk$, and $\varphi(\omega)$ the Fourier transform of $|\bk|/\|\bk\|_1$. Then
\begin{equation*}
  \erk(\bJ) = \frac{\big(\sum_\omega|\hat k(\omega)|^2\big)^2}{\sum_\omega|\hat k(\omega)|^4},
  \qquad
  \integ(\bJ) = 1-\max_{\omega\neq0}|\varphi(\omega)|.
\end{equation*}
Transparency is the participation ratio of the kernel's power spectrum: how evenly it passes spatial frequencies. Integrity measures how well the kernel mixes the slowest spatial mode. A wide blur on a small grid has low transparency and high integrity, the identity the reverse. For a Gaussian blur of width $\sigma$, $\integ\approx 1-\exp(-2\pi^2\sigma^2/N^2)$: it grows with the width of the kernel and \emph{falls} with the size of the grid. Local interaction over long distances is easy to cut in two. On Hypothesis~\ref{h:first}, a large system wired only locally is barely one system; this suggests that unity at scale needs long-range connections.

\subsection{Jacobian facets can resemble inputs}
\label{ssec:facets}

An MLP is trained to classify images into $n$ categories. When it is presented with an image from one of the categories, a dog say, how closely does the structure of its Jacobian resemble a dog?

Suppose the MLP uses rectified linear neurons:
\begin{equation*}
  \relu(a)=\max(0,a)
  \quad
  \relu'(a) = \begin{cases}
    1 & a > 0\\
    0 & a \leq 0
  \end{cases}
\end{equation*}
The net computes $\bz^{(l)} = \relu(\ba^{(l)})$ with $\ba^{(l)} = \bW_l \bz^{(l-1)}$ for each layer. Neuron $j$ in layer $l$ is \emph{active} if $a^{(l)}_j > 0$ and inactive otherwise. The final layer of the MLP produces $n$-dim output $\by$; its prediction is $\hat{c} := \argmax \by$, the coordinate with highest value.

Consider the input Jacobian. Each of the $n$ rows of $\bJ_{\by\leftarrow \bx}$, one per output neuron, is a \textbf{facet}. Each facet's similarity to the input is measured by the dot-product $s_i := \langle \bJ_{y_i\leftarrow \bx}, \bx\rangle$. Since output neuron $y_i$ corresponds to a category, facet $\bJ_{y_i\leftarrow \bx}$ should relate to \emph{that} category. Relu nets without bias terms have the property that their outputs are the dot-product of the gradient with the input \citep{srinivas2019}:
\begin{equation}
  \label{eq:relu_sim}
  y_i = \langle \bJ_{y_i\leftarrow \bx}, \bx\rangle =: s_i
\end{equation}
Facets are then easily interpretable. The similarity of facet $\bJ_{y_i\leftarrow \bx}$ to input $\bx$ \emph{is} the output $y_i$ for that category. The larger $y_i$, the larger the dot-product match between facet and input. In neuroscience terms a facet is the neuron's local linear receptive field, and Eq.~\eqref{eq:relu_sim} says its response is the match between input and receptive field. 

\begin{figure}[ht]
    \centering
    \resizebox{0.9\columnwidth}{!}{  \tikzset{
    actax/.style={black!50, line width=0.48pt, -latex},
    actfn/.style={teal!68!black, line width=1.1pt},
    actgr/.style={teal!36, line width=0.78pt, densely dashed},
    opendot/.style={draw=teal!36, fill=white, circle,
                    minimum size=4.5pt, inner sep=0pt, line width=0.65pt},
  }

  \begin{tikzpicture}

  \begin{scope}[x=0.70cm, y=2.4cm]

    \node[font=\normalsize\bfseries, anchor=south] at (0, 1.20)
      {Sigmoid};

    \draw[actax] (-3.7,0) -- (3.9,0);
    \draw[actax] (0,-0.08) -- (0,1.20);

    \foreach \t in {-3,-2,-1,1,2,3}
      \draw[black!40, line width=0.35pt] (\t, 0.022) -- (\t,-0.022);
    \node[font=\scriptsize, anchor=north] at ( 0, -0.03) {$0$};
    \node[font=\scriptsize, anchor=north] at (-3, -0.03) {$-3$};
    \node[font=\scriptsize, anchor=north] at ( 3, -0.03) {$3$};

    \draw[black!40, line width=0.35pt] (-0.07,0.5) -- (0.07,0.5);
    \draw[black!40, line width=0.35pt] (-0.07,1.0) -- (0.07,1.0);
    \node[font=\scriptsize, anchor=east] at (-0.10, 0.5) {$\tfrac{1}{2}$};
    \node[font=\scriptsize, anchor=east] at (-0.10, 1.0) {$1$};

    \draw[actfn, smooth, samples=80]
      plot[domain=-3.5:3.5] (\x, {1/(1+exp(-\x))});

    \draw[actgr, smooth, samples=80]
      plot[domain=-3.5:3.5] (\x, {exp(-\x)/(1+exp(-\x))^2});

    \node[font=\scriptsize, teal!75!black, anchor=west] at (1.8, 0.91) {$\sigma$};
    \node[font=\scriptsize, teal!50!black, anchor=west] at (0.4, 0.27) {$\sigma'$};

  \end{scope}

  \begin{scope}[xshift=6.4cm, x=0.70cm, y=0.80cm]

    \node[font=\normalsize\bfseries, anchor=south] at (0, 3.6)
      {ReLU};

    \draw[actax] (-3.7,0) -- (3.9,0);
    \draw[actax] (0,-0.25) -- (0,3.75);

    \foreach \t in {-3,-2,-1,1,2,3}
      \draw[black!40, line width=0.35pt] (\t, 0.07) -- (\t,-0.07);
    \node[font=\scriptsize, anchor=north] at ( 0, -0.13) {$0$};
    \node[font=\scriptsize, anchor=north] at (-3, -0.13) {$-3$};
    \node[font=\scriptsize, anchor=north] at ( 3, -0.13) {$3$};

    \foreach \u in {1,2,3}
      \draw[black!40, line width=0.35pt] (-0.20,\u) -- (0.20,\u);
    \foreach \u in {1,2,3}
      \node[font=\scriptsize, anchor=east] at (-0.30, \u) {$\u$};

    \draw[actfn] (-3.5,0) -- (0,0);
    \draw[actfn] (0,0) -- (3.5,3.5);

    \draw[actgr] (-3.5,0) -- (-0.06,0);
    \draw[actgr] ( 0.06,1) -- ( 3.5,1);
    \node[opendot] at (0,0) {};
    \node[opendot] at (0,1) {};

    \node[font=\scriptsize, teal!75!black, anchor=west] at (1.8, 2.55) {$\mathrm{ReLU}$};
    \node[font=\scriptsize, teal!50!black, anchor=west] at (1.8, 1.32) {$\mathrm{ReLU}'$};

  \end{scope}

  \end{tikzpicture}}
    \caption{Sigmoid and relu activations (solid) with their gradients (dashed).
      Sigmoid gradients vanish for large $|x|$; relu gradient is either zero (dead neuron) or constant.}
    \label{fig:activations}
  \figalttext[Sigmoid and relu activations with their gradients.]{Two plots. The sigmoid rises smoothly from 0 to 1 and its gradient is a bump that vanishes for large inputs. The relu is zero for negative inputs and linear for positive inputs, and its gradient is a step from 0 to 1.}%
\end{figure}

In short, when the image is a dog, and the net says it is a dog, then the dog-neuron's facet has large dot-product with the image. You may think this is trivial or tautological. It is not. It completely breaks for sigmoids.

\paragraph{Saturating sigmoids.}
For sigmoid neurons, the gradient approaches 0 as the output approaches 0 from above or 1 from below, see Fig.~\ref{fig:activations}. Eq.~\eqref{eq:relu_sim} does not hold for sigmoids; the clean relationship between facets and the original inputs breaks down.

One of the original justifications for switching from sigmoids to relus \citep{glorot2011} was that ``cortical neurons are rarely in their maximum saturation regime'' \citep{gutnick1995,douglas1995}. A situation where cortical neurons may saturate (analogous to a sigmoid approaching 1) is epileptic seizures, which are associated with loss of awareness \citep{balduzzi2008,blumenfeld2012}.

\subsection{Vivid and obscure errors}
\label{ssec:loss}

Descartes called a perception \emph{clear} when it is ``present and accessible to the attentive mind'' \citep{descartes1644}; here, \emph{vivid}. Operationally, define entries in a Jacobian, or groups of entries such as a facet or a pathway, as vivid when their magnitude is large relative to the rest of the Jacobian and obscure when it is near zero. Vividness is relative, so it is unchanged by rescaling the whole Jacobian. When are errors more or less vivid?

Attach a cross-entropy loss to the relu net above. Let $\bz = \text{softmax}(\by)$ and $\ell(\bz, c) = - \log(z_c)$ with gradient
\begin{equation*}
  e_i :=
  \frac{\partial\ell}{\partial y_i} = z_i - \mathbf{1}_{i=c}
  =\begin{cases}
    z_i - 1 & \text{if }i=c \\
    z_i & \text{else.}
  \end{cases}
\end{equation*}
The output is one-dimensional and the Jacobian collapses from $n\times d_l d_{l-1}$ to $1 \times d_l d_{l-1}$: $\bJ_{\ell\leftarrow \bW_l} = \mathbf{e}^\intercal\cdot \bJ_{\by\leftarrow \bW_l}$. There is nothing stopping us from taking \emph{both} the final layer $\by$ and the loss $\ell$ as outputs resulting in an $(n+1)\times d_l d_{l-1}$ Jacobian
\begin{equation*}
  \bJ_{(\by,\ell)\leftarrow \bW_l} =
  \left(\begin{array}{c}
    \bI_{n\times n} \\
    \mathbf{e}^\intercal
  \end{array}\right)
  \cdot \bJ_{\by\leftarrow \bW_l}
  = \left(\begin{array}{c}
    \bJ_{\by\leftarrow \bW_l} \\
    \mathbf{e}^\intercal\cdot \bJ_{\by\leftarrow \bW_l}
  \end{array}\right)
\end{equation*}
The rows are facets. The first $n$ describe how parameters influence the $n$ output neurons. The final $(n+1)^\textrm{st}$ row is an additional facet corresponding to the gradient of the loss. Large error-gradients make this facet vivid; as the net improves its performance, the facet fades into obscurity.

Biological brains may optimise many different cost functions, for example unsupervised objectives for predictive processing that models interoceptive and exteroceptive feedback \citep{seth2013}, as well as more clearly defined reward signals \citep{marblestone2016}. The relative salience or vividness of these components will depend on their sensitivity.

\subsection{The structure of attention}
\label{ssec:attention}

It is interesting to consider which pathways are vivid and obscure in a simplified model of attention:
\begin{equation}
  \label{eq:att}
  \softmax\big(\bX\bW_Q\cdot (\bX\bW_K)^\intercal\big)\cdot \bX\bW_{VO}
\end{equation}
where $\bX$ is $T\times d$. The softmax is applied to each row of $\bS := \bX\bW_Q\cdot (\bX\bW_K)^\intercal$ independently, and attention is causal: query $t$ attends to positions $s\le t$.

Inputs follow two paths through attention: \emph{(1)} the key-query (KQ) path, which is summarised as $\softmax(\bS)$, and \emph{(2)} the value-output (VO) path. The KQ path selects which tokens in the sequence to attend to, the VO path transforms and propagates what the KQ path selects. If the attention matrix is diffuse, then the layer attends to many tokens and gradients pass through the KQ path. If attention concentrates, for example in an attention sink \citep{fesser2026}, assigning $\approx1$ to one token and $\approx0$ to others, then the softmax derivative $\diag(\mathbf{p})-\mathbf{p}\mathbf{p}^\intercal$ at attention weights $\mathbf{p}$ approaches zero and gradients do not pass through the KQ path. The attention mechanism saturates; the decision what-to-attend-to is no longer \emph{live}; it has faded into obscurity. Gradients always propagate through the VO path. \emph{What is attended to} is vivid; \emph{what is not attended to} is obscure; \emph{why} tokens are or are not attended can be more or less vivid, and fades away with increased certainty, Fig.~\ref{fig:att}. A loose neural analogue is biased competition \citep{desimone1995}: once competition between stimuli is resolved, small changes to the losing stimuli no longer affect which one wins.

\begin{figure}[htbp]
  \centering
  \resizebox{0.9\columnwidth}{!}{\tikzset{
    att/.style={
      minimum size=1.45cm, inner sep=2pt, align=center, font=\small,
      draw=teal!65, line width=0.5pt
    },
    masked/.style={
      minimum size=1.45cm, inner sep=2pt, align=center, font=\small,
      fill=gray!6, draw=gray!28, line width=0.3pt, text=gray!45
    },
  }
  \def\B{1.45}
  \begin{tikzpicture}

  \begin{scope}

    \node[font=\normalsize\bfseries, anchor=south] at ({2*\B},{4*\B+0.4})
      {Diffuse attention};

    \node[att, fill=teal!70] at ({0.5*\B},{3.5*\B}) {$1.00$};
    \node[masked]            at ({1.5*\B},{3.5*\B}) {$0$};
    \node[masked]            at ({2.5*\B},{3.5*\B}) {$0$};
    \node[masked]            at ({3.5*\B},{3.5*\B}) {$0$};

    \node[att, fill=teal!35] at ({0.5*\B},{2.5*\B}) {$0.50$};
    \node[att, fill=teal!35] at ({1.5*\B},{2.5*\B}) {$0.50$};
    \node[masked]            at ({2.5*\B},{2.5*\B}) {$0$};
    \node[masked]            at ({3.5*\B},{2.5*\B}) {$0$};

    \node[att, fill=teal!23] at ({0.5*\B},{1.5*\B}) {$0.33$};
    \node[att, fill=teal!23] at ({1.5*\B},{1.5*\B}) {$0.33$};
    \node[att, fill=teal!23] at ({2.5*\B},{1.5*\B}) {$0.33$};
    \node[masked]            at ({3.5*\B},{1.5*\B}) {$0$};

    \node[att, fill=teal!18] at ({0.5*\B},{0.5*\B}) {$0.25$};
    \node[att, fill=teal!18] at ({1.5*\B},{0.5*\B}) {$0.25$};
    \node[att, fill=teal!18] at ({2.5*\B},{0.5*\B}) {$0.25$};
    \node[att, fill=teal!18] at ({3.5*\B},{0.5*\B}) {$0.25$};

    \draw[black!80, line width=1.1pt] (0,0) rectangle ({4*\B},{4*\B});
    \foreach \k in {1,2,3}{
      \draw[gray!35, densely dashed, line width=0.35pt]
        ({\k*\B},0)     -- ({\k*\B},{4*\B})
        (0,{\k*\B})     -- ({4*\B},{\k*\B});
    }
    \foreach \t in {1,...,4}
      \node[font=\small, anchor=east] at ({-0.12},{(4.5-\t)*\B}) {$q_\t$};
    \foreach \s in {1,...,4}
      \node[font=\small, anchor=north] at ({(\s-0.5)*\B},{-0.12}) {$k_\s$};
    \draw[decorate, decoration={brace, amplitude=5pt, mirror}]
      (0,{-0.65}) -- ({4*\B},{-0.65})
      node[midway, below=7pt, font=\footnotesize] {$T$};
    \draw[decorate, decoration={brace, amplitude=5pt}]
      ({-0.8},0) -- ({-0.8},{4*\B})
      node[midway, left=8pt, font=\footnotesize, rotate=90, anchor=south] {$T$};

  \end{scope}

  \begin{scope}[xshift=8.4cm]

    \node[font=\normalsize\bfseries, anchor=south] at ({2*\B},{4*\B+0.4})
      {Concentrated attention};

    \node[att, fill=teal!70] at ({0.5*\B},{3.5*\B}) {$1.00$};
    \node[masked]            at ({1.5*\B},{3.5*\B}) {$0$};
    \node[masked]            at ({2.5*\B},{3.5*\B}) {$0$};
    \node[masked]            at ({3.5*\B},{3.5*\B}) {$0$};

    \node[att, fill=teal!64] at ({0.5*\B},{2.5*\B}) {$0.92$};
    \node[att, fill=teal!6]  at ({1.5*\B},{2.5*\B}) {$0.08$};
    \node[masked]            at ({2.5*\B},{2.5*\B}) {$0$};
    \node[masked]            at ({3.5*\B},{2.5*\B}) {$0$};

    \node[att, fill=teal!3]  at ({0.5*\B},{1.5*\B}) {$0.04$};
    \node[att, fill=teal!6]  at ({1.5*\B},{1.5*\B}) {$0.09$};
    \node[att, fill=teal!61] at ({2.5*\B},{1.5*\B}) {$0.87$};
    \node[masked]            at ({3.5*\B},{1.5*\B}) {$0$};

    \node[att, fill=teal!2]  at ({0.5*\B},{0.5*\B}) {$0.03$};
    \node[att, fill=teal!62] at ({1.5*\B},{0.5*\B}) {$0.88$};
    \node[att, fill=teal!4]  at ({2.5*\B},{0.5*\B}) {$0.06$};
    \node[att, fill=teal!2]  at ({3.5*\B},{0.5*\B}) {$0.03$};

    \draw[black!80, line width=1.1pt] (0,0) rectangle ({4*\B},{4*\B});
    \foreach \k in {1,2,3}{
      \draw[gray!35, densely dashed, line width=0.35pt]
        ({\k*\B},0)     -- ({\k*\B},{4*\B})
        (0,{\k*\B})     -- ({4*\B},{\k*\B});
    }
    \foreach \t in {1,...,4}
      \node[font=\small, anchor=east] at ({-0.12},{(4.5-\t)*\B}) {$q_\t$};
    \foreach \s in {1,...,4}
      \node[font=\small, anchor=north] at ({(\s-0.5)*\B},{-0.12}) {$k_\s$};
    \draw[decorate, decoration={brace, amplitude=5pt, mirror}]
      (0,{-0.65}) -- ({4*\B},{-0.65})
      node[midway, below=7pt, font=\footnotesize] {$T$};

  \end{scope}
  \end{tikzpicture}}
  \caption{Attention matrices for a length-$T$ sequence.
    \textbf{Diffuse} (left): each query $q_t$ spreads weight uniformly over all past keys, giving equal attention $1/t$.
    \textbf{Concentrated} (right): each query peaks on a single past key, with most weight on one token.}
  \label{fig:att}
  \figalttext[Diffuse and concentrated causal attention matrices.]{Two lower-triangular attention matrices. Left: each query spreads its weight evenly over all earlier keys. Right: each query puts nearly all its weight on a single earlier key.}%
\end{figure}

\subsection{Shattered structure at initialisation}
\label{ssec:init}

James described the infant's world as ``one great blooming, buzzing confusion'' \citep{james1890}. Unfamiliar scenes are hard to parse because the relevant invariants are lacking; chess masters see and remember boards differently from novices \citep{chase1973}. To see how Jacobian structure differs between expert and novice, consider an extreme novice: a randomly initialised $\relu$ MLP. %

Following \citet{balduzzi:17shatter}, construct a relu net with 1D input and output, and 24 hidden layers with 200 relu neurons each, centred over the batch before each relu (the mean-subtraction step of batch normalisation), with the batch means held fixed. Compute gradients for inputs on a uniform linear grid of 256 points in the range $[-2,2]$. There is a 1D input Jacobian per point, and the question is how it varies across the grid. Nearby points --- that is, similar inputs --- should yield similar Jacobians, diverging as the distance between inputs increases.

\begin{figure}[t]
    \centering
    \begin{minipage}[t]{0.56\columnwidth}\vspace{0pt}
        \centering
        \panel{(a)}{\includegraphics[width=\linewidth]{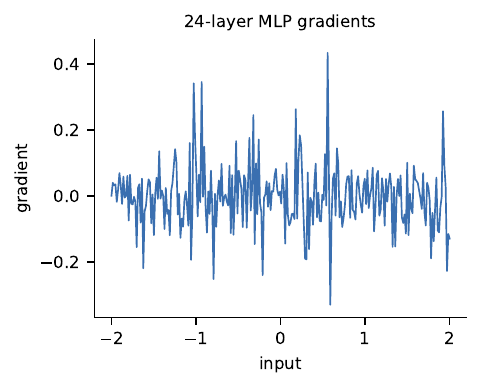}}
    \end{minipage}
    \hspace{0.03\columnwidth}%
    \begin{minipage}[t]{0.39\columnwidth}\vspace{0pt}
        \centering
        \panel{(b)}{\includegraphics[width=\linewidth]{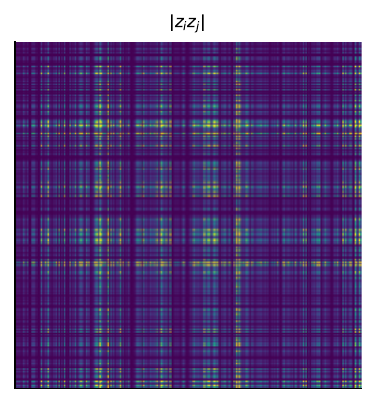}}
    \end{minipage}
    \caption{\textbf{Gradients in 24-layer relu MLP.}
    (a) 1-dim gradients plotted for inputs taken on uniform grid.
    (b) $|z_iz_j|$ for the gradients $z$ standardised across the grid.}
    \label{f:shatter}
  \figalttext[Gradients of a deep random relu network across a grid of inputs, and their products.]{Left: the gradient of a randomly initialised 24-layer relu network at 256 evenly spaced inputs looks like white noise. Right: the absolute products of these gradients, standardised across the grid, show no structure.}%
\end{figure}

Plotting the 256 1D gradients obtains a curve that is \emph{shattered}: that looks like white noise, Fig.~\ref{f:shatter}a. The matrix $|z_iz_j|$, for the gradients $z$ standardised across the grid, shows no structure, Fig.~\ref{f:shatter}b. This is because, on average, half of relu neurons are active for any given input. If it were the same half for every input, the gradients would all be identical and the curve would be a horizontal line. What actually happens is that the overlap between active neurons, for different inputs, decreases exponentially with depth, see \citet{balduzzi:17shatter}. Correlation between gradients is determined by the overlap, and so also decreases exponentially with depth. Across random nets, gradients at neighbouring grid points are almost perfectly correlated at depth $2$ and uncorrelated at depth $24$: the spatial structure of the grid is annihilated. For the randomly initialised MLP, the Jacobians at closely related inputs are wildly different. On Hypothesis~\ref{h:first}, this is the structural signature of a blooming buzzing confusion. 

Newborn babies likely lack the invariants required to perceive the typical motions of hands, mouths and bodies coherently, and the conjecture is that these are experienced as confusing, choppy jumbles. The MLP above is an extreme example of a system that lacks the invariants needed to coherently process its inputs, in this case the spatial structure of the grid.

\subsection{Distinct and confused facets}
\label{ssec:distinct}

Following Descartes, a perception is \emph{distinct} if, as well as being vivid, ``it is so sharply separated from all other perceptions'' \citep{descartes1644}. Humans can ``walk and chew gum at the same time'': hold multiple thoughts simultaneously, and moreover keep them separate. This is a learned skill; one cannot daydream when swimming for the first time. How does experience change such that you can, eventually, think and swim simultaneously? Intuitively, mastering swimming involves learning task-specific circuits that do not interfere with unrelated activities. Here is a more abstract formulation:
\begin{definition}
  \label{def:distinct}
  Given \textbf{parameter} Jacobian $\bJ_{n\times p}$, there are (at least) $k$ \textbf{distinct ideas} if there is a block diagonal permutation of the $n\times n$ coherency matrix $\bG := \bJ\bJ^\intercal$ with $k$ blocks.
\end{definition}
\noindent
Note a diagonal coherency matrix does not imply the Jacobian is block diagonal. Distinctness and integrity are different: a Jacobian with diagonal coherency matrix can have high integrity, \S\ref{app:integrity}.

The definition requires unpacking. Suppose the $n$ output neurons are of $k=2$ types, chewing and walking respectively. Two behaviours are distinct in this local sense if their parameter-sensitivity vectors are orthogonal: if how you chew gum is orthogonal to how you walk. Completely distinct ideas are unrealistic. It is more likely that \emph{near-orthogonal} cortical circuits arise. That is, approximate block structure with off-block-diagonal entries \emph{near zero} may arise.

It is useful to work out the details for an MLP, focusing on a particular layer. Pick outputs $\by$ and let $\bJ_{\by \leftarrow \bW_l}$ denote the $n\times d_l \times d_{l-1}$ Jacobian (a three-tensor) with respect to parameters in layer $l$. Then
\begin{equation*}
  \bJ_{\by \leftarrow \bW_l} = \nabla_{\bW_l}\by(\bx, \cW)
  = \bJ_{\by\leftarrow \ba_l} \otimes \bz_{l-1}^\intercal
\end{equation*}
where $\bJ_{\by\leftarrow \ba_l}$ is an $n \times d_l$ input Jacobian, but now the input refers to the pre-activations of layer $l$, which feed subsequent layers; not the input to the MLP. If $\bJ_{\by\leftarrow \ba_l}$ has SVD $\bU\bSigma\bV^\intercal$ then the SVD of $\bJ_{\by\leftarrow \bW_l}$ is
\begin{equation}
  \label{eq:facjac}
  \bU\cdot(\|\bz_{l-1}\|\cdot \bSigma)\cdot \left(\bV\otimes \frac{\bz_{l-1}}{\|\bz_{l-1}\|}\right)^\intercal.
\end{equation}
The left singular vectors of $\bJ_{\by\leftarrow \bW_l}$ only depend on $\bJ_{\by\leftarrow \ba_l}$, the Jacobian of the circuit connecting layer $l$'s activations to the output. So distinct ideas arise in the parameter Jacobian at layer $l$ when $\bJ_{\by\leftarrow \ba_l}\bJ_{\by\leftarrow \ba_l}^\intercal$ is block diagonal. That happens if activations in layer $l$ split into $k$ types that are processed separately downstream, due to gating, separation in the residual stream, or some other mechanism. In short, if the circuits do not interfere with each other, Fig.~\ref{fig:distinct_circuit}.

\begin{figure}[ht]
  \centering
  \resizebox{0.95\columnwidth}{!}{\tikzset{
  mixnode/.style={circle, minimum size=0.28cm, inner sep=0pt,
    fill=teal!22, draw=teal!55, line width=0.42pt},
  anode/.style={circle, minimum size=0.28cm, inner sep=0pt,
    fill=teal!62, draw=teal!82, line width=0.5pt},
  bnode/.style={circle, minimum size=0.28cm, inner sep=0pt,
    fill=teal!30, draw=teal!55, line width=0.5pt, dashed},
  mixedge/.style={gray!45, line width=0.32pt},
  aedge/.style={teal!68, line width=0.5pt},
  bedge/.style={teal!38, line width=0.5pt, dash pattern=on 2.2pt off 1.6pt},
  gcell/.style={minimum size=0.85cm, inner sep=1pt, align=center, font=\scriptsize,
    draw=teal!58, line width=0.42pt},
  gzero/.style={minimum size=0.85cm, inner sep=1pt, align=center, font=\scriptsize,
    fill=gray!7, draw=gray!25, line width=0.28pt, text=gray!40},
}

\def\dx{1.55}
\def\rA{2.7}
\def\rB{1.9}
\def\rC{1.1}
\def\rD{0.3}

\begin{tikzpicture}

\node[font=\normalsize\bfseries, anchor=south] at ({2*\dx},3.55)
  {Independent circuits after layer $l$};

\foreach \i in {\rA,\rB,\rC,\rD}{
  \foreach \j in {\rA,\rB,\rC,\rD}{
    \draw[mixedge] (0,\i) -- (\dx,\j);
    \draw[mixedge] (\dx,\i) -- (2*\dx,\j);
  }
}
\foreach \i in {\rA,\rB,\rC,\rD}{
  \node[mixnode] at (0,\i) {};
  \node[mixnode] at (\dx,\i) {};
  \node[mixnode] at (2*\dx,\i) {};
}
\node[font=\scriptsize, anchor=north] at (\dx,{\rD-0.35}) {mixed};

\foreach \i in {\rA,\rB}{
  \foreach \j in {\rA,\rB}{
    \draw[aedge] (2*\dx,\i) -- (3*\dx,\j);
  }
}
\foreach \i in {\rC,\rD}{
  \foreach \j in {\rC,\rD}{
    \draw[bedge] (2*\dx,\i) -- (3*\dx,\j);
  }
}
\foreach \i in {\rA,\rB}{
  \node[anode] at (3*\dx,\i) {};
}
\foreach \i in {\rC,\rD}{
  \node[bnode] at (3*\dx,\i) {};
}

\foreach \i in {\rA,\rB}{
  \foreach \j in {\rA,\rB}{
    \draw[aedge] (3*\dx,\i) -- (4*\dx,\j);
  }
}
\foreach \i in {\rC,\rD}{
  \foreach \j in {\rC,\rD}{
    \draw[bedge] (3*\dx,\i) -- (4*\dx,\j);
  }
}
\foreach \i in {\rA,\rB}{
  \node[anode] at (4*\dx,\i) {};
}
\foreach \i in {\rC,\rD}{
  \node[bnode] at (4*\dx,\i) {};
}

\node[font=\scriptsize, teal!75!black, anchor=south] at ({3.5*\dx},{\rA+0.42}) {circuit A};
\node[font=\scriptsize, teal!45!black, anchor=north] at ({3.5*\dx},{\rD-0.42}) {circuit B};

\draw[gray!50, densely dashed, line width=0.5pt]
  ({2*\dx+0.5},-0.55) -- ({2*\dx+0.5},3.15);
\node[font=\small, anchor=south] at ({2*\dx+0.5},3.2) {$\bm{a}_l$};

\node[font=\small, anchor=north] at (0,{\rD-0.35}) {$\bm{x}$};
\node[font=\small, anchor=north] at (4*\dx,{\rD-0.35}) {$\bm{y}$};

\begin{scope}[xshift=8.0cm, yshift=-0.55cm]

  \node[font=\small, anchor=south] at (1.7,4.15)
    {$\bm{G}=\bm{J}_{\bm{y}\leftarrow \bm{a}_l}\bm{J}_{\bm{y}\leftarrow \bm{a}_l}^\intercal$};

  \node[gcell, fill=teal!58] at (0.5,3.5) {$0.75$};
  \node[gcell, fill=teal!30] at (1.5,3.5) {$0.38$};
  \node[gzero] at (2.5,3.5) {$0$};
  \node[gzero] at (3.5,3.5) {$0$};

  \node[gcell, fill=teal!30] at (0.5,2.5) {$0.38$};
  \node[gcell, fill=teal!66] at (1.5,2.5) {$0.85$};
  \node[gzero] at (2.5,2.5) {$0$};
  \node[gzero] at (3.5,2.5) {$0$};

  \node[gzero] at (0.5,1.5) {$0$};
  \node[gzero] at (1.5,1.5) {$0$};
  \node[gcell, fill=teal!25] at (2.5,1.5) {$0.32$};
  \node[gcell, fill=teal!18] at (3.5,1.5) {$0.22$};

  \node[gzero] at (0.5,0.5) {$0$};
  \node[gzero] at (1.5,0.5) {$0$};
  \node[gcell, fill=teal!18] at (2.5,0.5) {$0.22$};
  \node[gcell, fill=teal!28] at (3.5,0.5) {$0.36$};

  \draw[teal!78, line width=0.95pt] (0,2) rectangle (2,4);
  \draw[teal!45, line width=0.95pt] (2,0) rectangle (4,2);

  \draw[black!80, line width=1.0pt] (0,0) rectangle (4,4);
  \foreach \k in {1,2,3}{
    \draw[gray!35, densely dashed, line width=0.32pt]
      (\k,0) -- (\k,4)
      (0,\k) -- (4,\k);
  }
  \node[font=\small, anchor=east] at (-0.12,3.5) {$y_{A1}$};
  \node[font=\small, anchor=east] at (-0.12,2.5) {$y_{A2}$};
  \node[font=\small, anchor=east] at (-0.12,1.5) {$y_{B1}$};
  \node[font=\small, anchor=east] at (-0.12,0.5) {$y_{B2}$};
  \foreach \s/\lab in {1/y_{A1}, 2/y_{A2}, 3/y_{B1}, 4/y_{B2}}
    \node[font=\small, anchor=north] at ({\s-0.5},-0.12) {$\lab$};

  \node[font=\footnotesize\itshape, teal!75!black] at (2,-0.65)
    {block diagonal $\Rightarrow$ distinct ideas};

\end{scope}

\end{tikzpicture}}
  \caption{An MLP whose activations at layer $l$ split into two circuits that do not interact again downstream.
    \textbf{Left:} before layer $l$ processing is mixed (grey); after layer $l$, circuit A (solid) and circuit B (dashed) evolve independently.
    \textbf{Right:} the coherency matrix $\bG=\bJ_{\by\leftarrow \ba_l}\bJ_{\by\leftarrow \ba_l}^\intercal$ is block diagonal, one block per circuit: two distinct ideas.}
  \label{fig:distinct_circuit}
  \figalttext[A network that splits into two noninteracting circuits, and its coherency matrix.]{Left: processing is mixed before layer l and then splits into two circuits that evolve independently. Right: the coherency matrix is block diagonal, one block per circuit.}%
\end{figure}

The opposite of distinct, confused, relates to neural superposition \citep{elhage2022}. Take $\relu(\bW^\intercal \bW \bx + \bb)$ and define features as independent, i.e. maximally distinct, if $\bW^\intercal \bW$ is diagonal, otherwise they are in \emph{superposition} (confused) because they interact. Note the analysis in \citet{elhage2022} depends on weights, not the Jacobian.

In mammalian brains, ideas may be kept distinct by shifting processing to the more modular circuitry in the cerebellum or basal ganglia \citep{alexander1986,middleton2000,graybiel2008,hull2022}, or in human brains by using the additional space provided by the enormously enlarged cerebral cortex.

\subsection{Learning: higher-order structure}
\label{ssec:learn}

Learning modifies the machinery of interaction itself. This section derives the Jacobians of weight updates, which yield a second-order experience.

Neural nets mostly learn by stochastic gradient descent (SGD). SGD computes the gradient of the loss on a random sample, then makes a small update of the weights in the opposite direction to the gradient, which typically slightly reduces the loss. Neural net training repeats this procedure many times.

Given objective function $f_\bw(\bx)$, the gradient of a batch of inputs $\{\bx_i\}_{i=1}^s$ is
\begin{equation*}
  \bg_t := \sum_{i=1}^s \nabla_\bw \big[f_\bw(\bx_i)\big]
  = \sum_{i=1}^s \bJ_{f_\bw(\bx_i) \leftarrow\bw}
\end{equation*}
Suppose neurons, on occasion, produce updated parameter vectors as an additional output of the system: $\bw_t = \bw_{t-1} - \eta_t \cdot \sum_i \bJ_{f_\bw(\bx_i) \leftarrow\bw}$. Since updating weights is itself a physical interaction, it has Jacobians of its own:
\begin{align*}
  \bJ_{\bw_t\leftarrow\bx}
  & = -\eta_t\cdot \nabla_\bx\sum_{i=1}^s \bJ_{f_\bw(\bx_i)\leftarrow\bw}
  \\
  \bJ_{\bw_t\leftarrow\bw}
  & = -\eta_t\cdot \nabla_\bw\sum_{i=1}^s \bJ_{f_\bw(\bx_i)\leftarrow \bw}
\end{align*}
differentiating only through $\bg_{t}$, and not through $\bw_{t-1}$, which is stop-gradiented for simplicity. The structure of learning is constituted by \textbf{second-order terms} like $\nabla^2_{\bw\bx} \left[f_\bw(\bx_i)\right]$ and $\nabla^2_{\bw} \left[f_\bw(\bx_i)\right]$.

Further, \emph{neural nets acquire knowledge by accumulating error gradients, a facet of their experience, into their weights.} This makes neural nets empiricists in a literal sense: what they know is their accumulated experience. They accumulate only the facets their training objective makes salient.

The forward and backward passes are distinct processes, with activations written to memory in one and read in the other. A batched forward pass diagonalises over the batch, one independent structure per sequence, while the backward pass aggregates errors across it: many parallel knots forward; another quite different knot backward. This is a quirk of how nets are trained today. In contrast, brains learn continually.

\subsection{Hardware}
\label{ssec:hardware}

Differential functionalism is committed to \emph{physical interactions}, so it matters how computation in data centres differs from the idealised model. What follows conjecturally applies Hypothesis~\ref{h:first} to hardware.

Digital computers use finite sets of numbers, floats and ints, over which $\lim_{h\to 0}$ is undefined. Strictly speaking, there are no gradients. Taking a hard line on Hypothesis~\ref{h:first} yields \textbf{floating point zombies}: computations over floats and over reals are practically indistinguishable, but one has experience and the other does not \citep{chalmers1996}. A more pragmatic alternative is to consider gradients as one window into interactions among many, finite differences being another. In this subsection, rather than introduce new measures specific to digital computation, I focus on where the existing measures collapse: leaving aside the discreteness of bits, what are the constraints on interactions in data centres relative to the idealised model?

\paragraph{Memory.}
Babbage divided the Analytical Engine into a \emph{store} (\textbf{memory}), which holds numbers, and a \emph{mill} (\textbf{execution}), which operates on them; the division lives on in registers and tensor cores. Retention is an identity map: each stored scalar is a persistent channel with integral rank $1$. Time does not stop when values are stored; the mill carries on: other values are loaded, mixed and written back. Time passes as the system as a whole interacts while the stored value is kept isolated. Alongside the active computation it forms a separate Jacobian block, giving zero joint integrity so it adds nothing to the system's present. When it is read, the route from write to read is transmissive through time, like a focused eye's segment through space, \secref{ssec:boundaries}. \textbf{Persistence without mixing is the signature of storage.} In a recurrent net the past is woven into the present at every step, \secref{ssec:time}; here memory is consulted only when read. The same holds for code and parameters, which do not interact until loaded into the mill.

The Extended Mind hypothesis brings the store-mill split to life as Otto, a man who relies on his notebook for recall because of a memory impairment \citep{clark1998}. Clark and Chalmers argue Otto's notebook is part of his memory because it plays a functionally equivalent role to biological memory. However, while Otto goes about his day, the notebook sits in his pocket and interacts with nothing. It only contributes to his experience when he reads or writes it. The rest of the time, its contents may as well not exist. \textbf{A stored-program computer is like Otto, with notebooks for both its memories \emph{and code}.} Its experience is limited to what is happening in the mill. 

\paragraph{Tiles.}
A matmul in the idealised model is a single transformation. In a data centre it is thousands of tile operations, each mixing a small block, with partial results parked in memory. Over a window spanning the whole matmul the two agree; over shorter windows the data centre's Jacobian is block diagonal and its integrity is zero. Structure knits together only at coarse timescales, out of pieces that interact briefly. On this account, hardware \emph{strobes} experience relative to the idealised model.

\begin{figure}[htbp]
  \centering
  \resizebox{0.95\columnwidth}{!}{\begin{tikzpicture}[x=1cm,y=1cm]
\tikzset{nd/.style={circle,draw=tealmid,fill=teallt!40,line width=0.8pt,minimum size=4.4mm,inner sep=0pt}}
\begin{scope}
 \node[font=\scriptsize,anchor=south] at (2.1,2.65) {Idealised model};
 \foreach \i in {0,1,2,3}{\coordinate (a\i) at (0,{2.1-\i*0.7});\coordinate (b\i) at (4.2,{2.1-\i*0.7});}
 \foreach \i in {0,1,2,3}{\foreach \j in {0,1,2,3}{\draw[draw=teallt,line width=0.7pt] (a\i) -- (b\j);}}
 \draw[draw=teal,line width=0.9pt,dash pattern=on 4pt off 3pt,rounded corners=3pt] (1.2,-0.35) rectangle (3.0,2.45);
 \foreach \i in {0,1,2,3}{\node[nd] at (a\i) {};\node[nd] at (b\i) {};}
 \node[font=\scriptsize,text=teal,anchor=north] at (2.1,-0.5) {one mixing};
 \node[font=\scriptsize,text=teal,anchor=north] at (2.1,-0.95) {$\integ>0$ throughout};
 \node[font=\scriptsize\itshape,text=black!60,anchor=north] at (2.1,-1.35) {inputs are woven into outputs};
\end{scope}
\begin{scope}[xshift=6.2cm]
 \node[font=\scriptsize,anchor=south] at (2.4,2.65) {Datacenter};
 \foreach \t in {0,1,2}{
  \pgfmathsetmacro\x{\t*1.7}
  \draw[draw=tealmid,fill=teallt!14,line width=0.8pt,rounded corners=2pt] (\x,1.1) rectangle (\x+1.25,2.35);
  \foreach \i in {0,1,2}{\foreach \j in {0,1,2}{
    \draw[draw=teallt,line width=0.5pt] (\x+0.14,{1.28+\i*0.37}) -- (\x+1.11,{1.28+\j*0.37});}}
  \draw[draw=crimson!65,fill=crimson!7,line width=0.8pt,rounded corners=1.5pt] (\x,0.1) rectangle (\x+1.25,0.65);
  \draw[draw=crimson!80,line width=1.1pt] (\x+0.14,0.375) -- (\x+1.11,0.375);
  \draw[draw=crimson!45,line width=0.8pt,dash pattern=on 2.5pt off 2pt] (\x+0.625,1.1) -- (\x+0.625,0.65);
 }
 \node[font=\scriptsize,text=tealmid,anchor=west] at (0,-0.5) {execution: $\integ>0$};
 \node[font=\scriptsize,text=crimson,anchor=west] at (0,-0.95) {memory: $\integ=0$};
 \node[font=\scriptsize\itshape,text=black!60,anchor=north] at (2.4,-1.35) {small blocks, separated by inert memory};
\end{scope}
\end{tikzpicture}}
  \caption{\textbf{Tiles.} A matmul in the idealised model is one transformation; in a data centre it is thousands of tile operations separated by transfers through memory. The composite Jacobians agree; over shorter time windows the data centre's is block diagonal.}
  \label{fig:tiles}
  \figalttext[A dense matrix multiply compared with tiled computation in a data centre.]{Left: in the idealised model one matrix multiply mixes all inputs into all outputs at once. Right: in a data centre the same product is computed as many small tile operations separated by transfers through memory, so over short windows the computation splits into separate blocks.}%
\end{figure}

\paragraph{Data centres simulate the idealised model.}
They do so with three degradations: \emph{consultation}, as data and code are stored rather than interwoven; \emph{strobing}, as mixing knits together only at coarse timescales; and \emph{sharding}, across devices, whose Jacobians are block diagonal except when the devices exchange data. A person looking at a flashing photograph has a perfectly good visual system; the degradation is in the input. For neural nets in data centres it goes \emph{all the way in}: the flicker is in the perceiving, not in what is perceived.

\section{Discussion and conclusions}
\label{sec:conclusion}

The paper proposes \emph{differential functionalism}: the first-order structure of physical interactions, captured by Jacobians, shapes the structure of phenomenal experience. The hypothesis has two clauses: experience registers differences to the extent that they make a difference, and extends to the extent that interactions connect. Transparency and integrity quantify the two clauses; their zero sets are forced by the hypothesis.

Applied to neural networks, the measures behave sensibly. Integrity separates overlapping experiences (shared inputs) from merged systems (shared parameters). Computing integral rank for a perceiving system shows how it can select its boundary by focus, how the boundary comes closer when inputs are noisy (fog), and how an extended present arises from mixing across time over a spread of timescales, but not in an MLP applied frame by frame. The closed forms for convolution suggest that unity at scale needs long-range connections. Beyond the two measures, the structure of Jacobians separates the vivid from the obscure, in error signals and in attention, and the distinct from the confused, and gives a structural description of a system that lacks the invariants of its world. Learning adds second-order structure, \secref{ssec:learn}. In brains, the measures can be estimated from stimulation data, \secref{app:tests}.

The same computation can have different structure on different substrates and, because differential functionalism is committed to physical interactions, the differences matter. If the structure of experience is shaped by the structure of physical interactions, then whatever experience a neural net has when it \emph{runs} is, conjecturally, limited by its hardware implementation, \secref{ssec:hardware}: stored memories and parameters contribute nothing between reads, and tiled computation knits together only over coarse windows. By contrast, for computational functionalism only the abstract computation matters \citep{putnam1975}; its canonical model, the Turing machine, is the storage--execution division in its purest form, as tape and head. Yet substrates running the same computation differ in speed, energy and which algorithms they make practical, enough to steer the direction of AI research \citep{hooker2020}.

\paragraph{Limitations.}
The analysis is first order and local: Jacobians describe responses to small perturbations of the actual state. Studying how responses vary requires comparing Jacobians at different states, as \secref{ssec:init} does across an input grid. Jacobians track perturbations, not the values perturbed; values enter only through the state at which the Jacobian is evaluated. 

The measures are model dependent, and comparisons are meaningful only for comparable models. The worked examples are chosen so that the analysis is interpretable; how the picture changes for more realistic models, and above all for brains, is open. Finally, Hypothesis~\ref{h:first} is silent on valence; a companion paper proposes that valence is the rate of change of the Jacobian's log scale due to interoception \citep{balduzzi2026v}.

\paragraph{Conclusion.}
Differential functionalism provides a candidate bridge between the structure of differentiable physical interactions and the structure of experience. As questions about consciousness in AI become pressing \citep{aru2023,butlin2023,chalmers2023}, it offers a way to ask how a system's experience, if it has any, would be structured.

\ifnum\arxivversion=0
\section*{Conflicts of interest}
\pending{to be added}

\section*{Funding}
\pending{to be added}

\fi

\section*{Code availability}
Code is available at \url{https://github.com/dbalduzzi/diff_func}.

\ifnum\arxivversion=0
\section*{Author contributions statement}
\pending{to be added}

\section*{Acknowledgements}
\pending{to be added}
\fi

\bibliographystyle{oup-abbrvnat}
\bibliography{refs}

\begin{appendices}

\section{Appendices}
\label{app}

\subsection{Singular value decomposition}
\label{app:svd}

Given $n \times d$ matrix $\mathbf A$, the \textbf{singular value decomposition (SVD)} is a factorisation \citep{stewart1993,golub2013,strang2019}
\begin{equation}
  \label{eq:svd}
  {\mathbf A} = {\mathbf U}{\mathbf \Sigma}{\mathbf V}^\intercal
\end{equation}
where ${\mathbf U}$ and ${\mathbf V}$ are orthogonal, $n \times n$ and $d \times d$
respectively, and $\mathbf{\Sigma}$ is an $n \times d$ matrix with diagonal entries $\sigma_i \geq 0$ for all $i$, and all off-diagonal entries equal to zero.

The diagonal entries $\sigma_i$ of $\mathbf{\Sigma}$ are
the \emph{singular values} of $\mathbf{A}$. Singular values are ordered so that $\sigma_1 \geq \sigma_2 \geq \cdots\geq \sigma_{\min(n, d)}$. The rank $\rk(\mathbf{A})$, typically denoted $r$, is the number of \emph{nonzero} singular values.

Geometrically, the SVD says every matrix factorises into a rotation followed by rescaling and another rotation. It follows that
\begin{equation*}
  \mathbf{A} \mathbf{v}_i = \sigma_i \mathbf{u}_i
  \quad\text{and}\quad
  \mathbf{A}^\intercal \mathbf{u}_i = \sigma_i \mathbf{v}_i
\end{equation*}
where $\mathbf{u}_i$ and $\mathbf{v}_i$ are orthonormal vectors in $\mathbb {R}^n$ and $\mathbb{R}^d$ respectively, given by the columns of $\mathbf{U}$ and $\mathbf{V}$.

\subsection{Effective rank}
\label{app:erk}

Effective rank, Definition~\ref{def:erk}, is also called the \textbf{participation ratio}. Its reciprocal, the inverse participation ratio, is the standard diagnostic for Anderson localisation in condensed matter physics \citep{wegner1980}: it distinguishes an eigenstate spread over the whole system from one concentrated on a few sites.

Litwin-Kumar et al.\ define the \textbf{dimension of a neural representation} as the participation ratio of the population covariance spectrum (they work with eigenvalues of covariance matrices, which are squared singular values of the factor matrix). They show it is maximised at degrees of synaptic connectivity that match those empirically observed in cerebellar granule cells and \emph{Drosophila} Kenyon cells \citep{litwin-kumar2017}.

For the remainder of this subsection, suppose matrix $\bA$ has at least one nonzero singular value. Here are some alternate perspectives on effective rank:
\begin{enumerate}
  \item \emph{Ratio of traces.}
  Effective rank can be rewritten as
  \begin{equation*}
    \erk(\bA) = \frac{\tr(\bA \bA^\intercal)^2}{\|\bA\bA^\intercal\|^2_F}
    =  \frac{\tr(\bA \bA^\intercal)^2}{\tr\big((\bA \bA^\intercal)^2\big)}
  \end{equation*}
  $\bA\bA^\intercal$ can be replaced with $\bA^\intercal\bA$ in the above without changing the result.

  \item \emph{Inverse collision probability.}
  If the squared singular values are normalised to form a probability distribution,
  \begin{equation*}
    p_i = \frac{\sigma_i^2}{\sum_{j=1}^r \sigma_j^2}
  \end{equation*}
  then effective rank is
  \begin{equation*}
    \erk(\mathbf{A}) = \frac{1}{\sum_{i=1}^r p_i^2},
  \end{equation*}
  where the denominator $\sum_{i=1}^r p_i^2$ is the \emph{collision probability}, the probability that two independent draws from distribution $p$ yield the same index.

  The collision probability varies between maximum of $1$, when there is one nonzero singular value, and minimum of $\frac{1}{r}$, when there are $r$ equal singular values. It follows that, when rank is nonzero, effective rank lies in $[1,r]$.

  \item \emph{Coefficient of variation of squared singular values.}
  If $\mu = \frac{1}{r}\sum_{i=1}^r\sigma_i^2$ and $s^2 = \frac{1}{r}\sum_{i=1}^r (\sigma_i^2-\mu)^2$ are the mean and variance of the squared singular values respectively, then effective rank is
  \begin{equation*}
    \erk(\bA) = \frac{r}{1 + (s/\mu)^2}
    \quad\text{ when }\rk(\bA)>0,
  \end{equation*}
  where $\frac{s}{\mu}$ is the \emph{coefficient of variation} of the squared singular values.

  By this formula, effective rank is rank (the numerator) penalised by unevenness of the squared singular values. If the variance of squared singular values is zero, then effective rank is the rank. As variance increases, effective rank decreases.
\end{enumerate}

\subsection{Integrity}
\label{app:integrity}

\paragraph{Laplacians.}
Given a graph with symmetric edge weights $w_{ij}\ge0$, its Laplacian is $L_{ij}=-w_{ij}$ for $i\neq j$ and $L_{ii}=\sum_{k\neq i}w_{ik}$ \citep{harary1967,bollobas1998,stanley1999}. For a graph with at least two vertices and positive degrees, with $\bD$ the diagonal of degrees, the normalised Laplacian is $\mathcal{L}=\bD^{-1/2}\bL\bD^{-1/2}$. Its eigenvalues lie in $[0,2]$, the smallest is $0$, and the second smallest is $0$ iff the graph is disconnected \citep{chung1997}.

\paragraph{Spectrum.}
For the nonempty active bipartite graph $G_*(\bA)$ of Definition~\ref{def:integrity}, write $n=n_*$ and $d=d_*$. Then $\mathcal{L}=\bI-\big(\begin{smallmatrix}\bzero&\calA\\ \calA^\intercal&\bzero\end{smallmatrix}\big)$. Its eigenvalues are $1\pm\sigma_i(\calA)$, together with $|n-d|$ eigenvalues equal to $1$. The square-rooted degrees form a singular pair of $\calA$ with $\sigma_1(\calA)=1$, giving the zero eigenvalue. When $\min(n,d)>1$, the next eigenvalue is $1-\sigma_2(\calA)=\integ(\bA)\in[0,1]$. When $\min(n,d)=1$ and $n+d>2$, it is $1$; for $n=d=1$, it is $2$ and integrity is instead set to $1$. Integrity equals $1$ iff $\sigma_2(\calA)=0$, i.e.\ iff $\calA$, equivalently $|\bA|$, has rank one.

\paragraph{Distinct ideas are not disconnected.}
Hadamard matrices have maximal integrity, yet their coherency matrices are diagonal because their rows are orthogonal. Alternatively, compute the singular value decomposition $\bA=\bU\bSigma\bV^\intercal$ of a random matrix and set $\bB:=\bSigma\bV^\intercal$. The coherency matrix $\bB\bB^\intercal=\bSigma^2$ is diagonal, but in general $\bB$ will have nonzero integrity. 

\paragraph{Interacting systems.}
Two MLPs, 64--32--32--10, are trained separately to $100\%$ training accuracy on the $8\times8$ scikit-learn digits. In the two-way case, each second hidden layer receives $\epsilon$ times a fixed random projection of the other's first hidden layer; in the one-way case, only one does. Integral rank is computed for the parameter Jacobian of the $20$ joint logits and compared with the larger of the two nets' own, for ten pairs of nets (training seeds and projections) at ten inputs each. With two-way coupling the nets are never one unit at $\epsilon\le1$; they are one in $5$ of $100$ cases at $\epsilon=1.5$, in $96$ at $\epsilon=2$ and in all $100$ at $\epsilon=3$. With one-way coupling they remain two units in every case, for every $\epsilon$ up to $3$. 

\subsection{Optics of boundaries}
\label{app:optics}
 
This appendix supplies the model behind \secref{ssec:boundaries}.
 
\paragraph{The model.}
Air alone does not transmit: each point of a tree illuminates the whole pupil, so without a lens every photoreceptor would sum light from every point, $\bJ_{\textrm{retina}\leftarrow\textrm{tree}}\propto\mathbf{1}\mathbf{1}^\intercal$. With a lens of power $P$, a ray from an object at distance $d$ lands on a retina at depth $z$ at a position independent of its direction iff $P=1/d+1/z$, the lens equation \citep{hecht2017}. Otherwise light from one point spreads over $z\,a\,|\Delta P|$ on the retina, for pupil diameter $a$ and focus error $\Delta P$ in dioptres. Neither the lens nor the photoreceptors mix on their own; mixing arises from their composition.
 
The scene is a signpost at $2\,$m and a tree at $50\,$m, $n=16$ points each, with one photoreceptor per point. The segment is $\bS=\diag(\bS_{\textrm{sign}},\bS_{\textrm{tree}})$. The stage after the retina is a dense Gaussian matrix $\bD$, $64\times32$; results are averaged over $100$ draws. Definition~\ref{def:boundary} compares the integral rank of the composite $\bD\bS$ with that of $\bD$ alone. The cut moves through a whole object's segment at once: replacing a single photoreceptor by the points that feed it is not a valid cut when the segment mixes, since those points also reach the outputs through neighbouring photoreceptors.
 
\paragraph{Focus.}
Defocus blurs each point with a Gaussian of width $c\,a\,|\Delta P|$ photoreceptor spacings, for pupil diameter $a$ and focus error $\Delta P$ in dioptres, with stylised $c=2$ per mm per dioptre. With a $4\,$mm pupil focused on the signpost, the tree is blurred over $3.8$ photoreceptors. Selection works in dioptres, not metres: focused at infinity, the tree's blur is $0.16$, and the tree transmits for any focus beyond about $16\,$m. An in-focus segment leaves integral rank exactly unchanged. With the tree as the only input, the out-of-focus tree lowers it to $0.13$ of its value on average, and in every draw; with the signpost in focus alongside, to $0.62$. With both objects as input, the ratio is at most $0.62$ over all foci.
 
\paragraph{Fog.}
A fraction $\theta=e^{-\beta d}$ of the light arrives unscattered, with $\beta=3.912/V$ for visibility $V$ \citep{koschmieder1924}. The rest spreads over the object: $\bS_{\textrm{tree}}=\theta\bI+(1-\theta)\mathbf{1}\mathbf{1}^\intercal/n$, which passes mean luminance and multiplies contrast by $\theta$. No focus can help. A downstream contrast gain $g$ multiplies contrast by $g\theta$. With photoreceptor noise $\sigma$, the optimal (Wiener) gain is $g=\theta/(\theta^2+\sigma^2)$, leaving contrast factor $\gamma_u=\theta^2/(\theta^2+\sigma^2)$. This stays near $1$ until
\begin{equation*}
  d^* = \frac{V\ln(1/\sigma)}{3.912}\approx0.77\,V \quad(\sigma=0.05),
\end{equation*}
beyond which the gain gives up and the tree is left mixed, Table~\ref{tab:fog}.
 
\begin{table}[t]
  \centering\small
  \caption{Fog pulls the boundary in. Visibility $100\,$m, $\sigma=0.05$, $d^*=77\,$m. Integral rank of the composite, tree plus in-focus signpost, relative to the boundary at the retina.}
  \label{tab:fog}
  \begin{tabular}{@{}rccccc@{}}
    \toprule
    & & & & \multicolumn{2}{c}{$\integ_{\erk}$ ratio} \\
    \cline{5-6}
    $d$ (m) & $\theta$ & gain & $\gamma_u$ & receptors & after gain\\
    \midrule
    10  & 0.68  & 1.5  & 0.995 & 0.91 & 1.00\\
    50  & 0.14  & 6.3  & 0.89  & 0.61 & 0.99\\
    77  & 0.05  & 10.0 & 0.49  & 0.56 & 0.80\\
    100 & 0.02  & 6.9  & 0.14  & 0.55 & 0.61\\
    150 & 0.003 & 1.1  & 0.003 & 0.55 & 0.55\\
    \botrule
  \end{tabular}
\end{table}
 
\paragraph{The effect of blur.}
Integral rank is $\erk\cdot\integ$ with $\integ\le1$, so blur can raise integrity by at most a factor $1/\integ(\bD)$. A dense $\bD$ already has integrity of about $0.75$, so local blur has little to add and costs transparency: with one object as input, mean integral rank falls from $9.4$ unblurred to $2.1$ at blur width $2$ and $1.1$ at width $4$. With a diagonal $\bD$, here the identity, nothing else mixes, integrity starts at $0$, and integral rank rises with blur, from $0$ to $0.27$ at width $2$ and $0.58$ at width $4$.

\paragraph{Robustness.}
The blur scale $c$ is stylised. For $c=1$ and $c=4$ the out-of-focus tree retains $0.24$ and $0.09$ of integral rank, and with both objects as input no focus scores above $0.70$. Only the depth of field changes: the tree transmits for focus beyond $9\,$m and $24\,$m respectively.
 
\paragraph{Limits.}
The model is deliberately minimal: one-dimensional objects, Gaussian defocus and a random downstream stage. Real defocus destroys some spatial frequencies, so it cannot be undone downstream and must be corrected optically.

\subsection{Closed form measures for convolutions}
\label{app:conv}

Take single-channel circular convolution on an $N\times N$ grid. 

\paragraph{Transparency.}
The Jacobian is doubly block-circulant, so the 2D Fourier transform diagonalises it with eigenvalues $\hat k(\omega)$; its singular values are $|\hat k(\omega)|$, which gives $\erk$. 

\paragraph{Integrity.}
On the periodic $N\times N$ grid, $|\bJ|$ is circulant with symbol $\|\bk\|_1\,\varphi(\omega)$. Every row and column of $|\bJ|$ sums to $\|\bk\|_1$, so $\calA=|\bJ|/\|\bk\|_1$, whose singular values are $|\varphi(\omega)|$. The largest is $\varphi(0)=1$, so
\begin{equation*}
  \integ(\bJ)=1-\max_{\omega\neq0}|\varphi(\omega)|.
\end{equation*}
For a Gaussian kernel of width $\sigma$, $\varphi(\omega)\approx\exp(-\sigma^2|\omega|^2/2)$ and the slowest nonzero mode has $|\omega|=2\pi/N$, giving $\integ\approx1-\exp(-2\pi^2\sigma^2/N^2)$. Numerically, for $\sigma=2$ and an untruncated periodic kernel, integrity is $0.27$ at $N=16$ and $0.07$ at $N=32$, equal to $1-\max_{\omega\neq0}|\varphi(\omega)|$ and within $0.001$ of the Gaussian approximation.

\subsection{Estimating Jacobians in brains}
\label{app:tests}
 
Jacobians can be estimated for brains using methods from perturbational neuroscience. Stimulating at one site and recording at many gives a column of a response matrix; stimulating at many sites gives the matrix. Cortico-cortical evoked potentials from intracranial stimulation \citep{matsumoto2004,keller2014} and source-localised TMS-EEG \citep{massimini2005,casali2013} provide response matrices, $\bR_{ij}(\tau)$, the response at site $i$ a latency $\tau$ after a small pulse at site $j$. To first order in the pulse amplitude, $\bR(\tau)$ is the Jacobian of the brain's dynamics from the stimulated to the recorded sites over a window $\tau$. Effective-connectivity models fitted to evoked or spontaneous activity estimate the same object indirectly, as the Jacobian of linearised dynamics \citep{friston2003,gilson2016}. Estimates are noisy, and noise fills in the zeros that make integrity vanish, so integrity must be compared with a noise floor, for example from sham stimulation.
  
Predictions:

\begingroup\renewcommand{\labelenumi}{P\arabic{enumi}.}
\begin{enumerate}
  \item \emph{Duration.} 
  The latency range over which $\bR(\tau)$ from early stimulation to late response keeps integrity above a noise floor, the duration of \secref{ssec:time}, should shrink in NREM sleep and anaesthesia. NREM responses are strong but brief and local, which predicts low integrity across $\tau$ even where early responses are large.
  \item \emph{Dissociations.} 
  The two measures come apart in characteristic ways, which gives predictions that a single complexity index cannot make. A generalised seizure, with activity locked to one pattern, should show high integrity and low transparency: every site responds, but along one direction. Fragmented states with intact local processing, such as NREM sleep, should show the reverse. Ketamine anaesthesia, where PCI remains high but responsiveness is lost \citep{sarasso2015}, is a natural test of whether transparency and integrity track PCI or not.
  \item \emph{Consistency with IIT.} 
  The measures should reproduce the pattern already found for the perturbational complexity index \citep{casali2013} and predicted by IIT: lower integrity and integral rank in dreamless NREM sleep and propofol anaesthesia than in wakefulness, and integrity falling to the noise floor across a cut corpus callosum. They would do so for the same reason, the loss of interaction.
\end{enumerate}
\endgroup

\paragraph{Artificial systems.}
The same measures can be computed exactly for neural networks. There they give a testbed for the framework's internal consistency: how transparency and integrity change over training, across layers, and at the representations identified as a global workspace in \citet{gurnee2026}. A local maximum of integral rank, in the sense of Definition~\ref{def:boundary}, specify a workspace.

\end{appendices}

\end{document}